\documentclass[sigconf,natbib=true, nonacm]{acmart}

\usepackage{enumitem}
\usepackage{relsize}
\usepackage{booktabs}
\usepackage{tabularx}
\usepackage{multirow}
\usepackage{graphicx}
\usepackage{xspace}
\usepackage{xcolor}
\usepackage{subcaption}
\usepackage{hyperref}

\usepackage{mathtools}
\usepackage{pifont}

\usepackage{tablefootnote}

\usepackage{cleveref}

\usepackage{tikz}
\usetikzlibrary{arrows.meta}
\usetikzlibrary{shapes.arrows}
\usetikzlibrary{shapes.misc}
\usetikzlibrary{fit}
\usetikzlibrary{calc,shapes}
\usetikzlibrary{tikzmark}
\usetikzlibrary{arrows}

\crefname{chapter}{Chapter}{}
\crefname{section}{Section}{Sections}
\crefformat{section}{Section~#2#1#3}

\crefname{table}{Table}{Tables}
\crefname{figure}{Figure}{Figures}
\crefname{algorithm}{Alg.}{Algs.}
\crefname{line}{Line}{Lines}
\crefname{appendix}{App.}{}
\crefname{chapter}{Chapter}{Chapters}

\crefname{thm}{Theorem}{Theorems}
\crefname{prop}{Proposition}{Propositions}
\crefname{definition}{Definition}{Definitions}
\crefname{lemma}{Lemma}{Lemmas}
\crefname{cor}{Corollary}{Corollaries}
\crefname{equation}{Eq.}{Eqs.}

\creflabelformat{equation}{#2\textup{#1}#3}
\crefrangelabelformat{equation}{(#3#1#4--#5#2#6)}
\creflabelformat{figure}{#2\textup{#1}#3}
\crefrangelabelformat{figure}{(#3#1#4--#5#2#6)}

\newcolumntype{C}{>{\centering\arraybackslash}X}

\renewcommand{\ldots}{\ensuremath{{\ldotp\kern-0.2em\ldotp\kern-0.2em\ldotp}}}

\renewcommand{\cdots}{\ensuremath{{\cdotp\kern-0.2em\cdotp\kern-0.2em\cdotp}}}
\renewcommand{\dots}{\ensuremath{{\ldotp\kern-0.2em\ldotp\kern-0.2em\ldotp}}}

\newcommand{\defn}[1]{\textbf{#1}}

\newcommand{\cmark}{\ding{51}}
\newcommand{\xmark}{\ding{55}}

\newcommand{\sota}{SotA\xspace}

\newcommand{\datasetName}{BuDDIE\xspace} 

\renewcommand{\paragraph}[1]{\vspace{5pt}\noindent\textbf{#1.}}

\allowdisplaybreaks

\begin{document}
\title{\datasetName: A Business Document Dataset for Multi-task Information Extraction}

\author[Zmigrod et al.]{Ran Zmigrod$^*$ $\quad$ Dongsheng Wang$^*$ $\quad$ Mathieu Sibue $\quad$ Yulong Pei $\quad$ Petr Babkin $\quad$ Ivan Brugere $\quad$ Xiaomo Liu $\quad$ Nacho Navarro $\quad$ Antony Papadimitriou $\quad$ William Watson $\quad$ Zhiqiang Ma $\quad$ Armineh Nourbakhsh $\quad$ Sameena Shah}
\email{{first\_name}.{last\_name}@jpmchase.com}
\affiliation{%
  \institution{J.P. Morgan AI Research}
  \city{}
  \country{}
}
%
\begin{abstract}
The field of visually rich document understanding (VRDU) aims to solve a multitude of well-researched NLP tasks in a multi-modal domain.
Several datasets exist for research on specific tasks of VRDU such as document classification (DC), 
key entity extraction (KEE), entity linking, visual question answering (VQA), \emph{inter alia}.
These datasets cover documents like invoices and receipts with sparse annotations such that they support one or two co-related tasks (e.g., entity extraction and entity linking).
Unfortunately, only focusing on a single specific of documents or task is not representative of how documents often need to be processed in the wild -- where variety in style and requirements is expected.
In this paper, we introduce \defn{\datasetName} (\textbf{Bu}siness \textbf{D}ocument \textbf{D}ataset for \textbf{I}nformation \textbf{E}xtraction), the first multi-task dataset of $1,\!665$ real-world business documents that contains rich and dense annotations for
DC, KEE, and VQA.
Our dataset consists of publicly available business entity documents from US state government websites.
The documents are structured and vary in their style and layout across states and types (e.g., forms, certificates, reports, etc.).
We provide data variety and quality metrics for \datasetName as well as a series of baselines for each task.
Our baselines cover traditional textual, multi-modal, and large language model approaches to VRDU.
\end{abstract}

\maketitle
\def\thefootnote{*}\footnotetext{Equal contribution.}
\def\thefootnote{\arabic{footnote}}
\section{Introduction}
Document images are ubiquitous in the real world; for example, reports, receipts, forms, certificates, \emph{inter alia} are integral throughout the business pipeline.
Modern systems need to efficiently and accurately capture and understand information from digital and scanned documents.
As a result, researchers from computer vision, machine learning, and NLP have focused on creating models for visually rich document understanding \cite{xu2020layoutlm,appalaraju2021docformer,davis2021visual,xu2021layoutlmv2,zhang2022multimodal}.
With rising interest in the field, the necessity for publicly available, large, and robust datasets is becoming ever-more evident.

Numerous datasets have been created to support the modeling of document understanding tasks such as DC, KEE, entity linking, and VQA \cite{jaume2019funsd,park2019cord,stanislawek-2021-kleister,mathew2021docvqa}.
Datasets often contain ground-truth annotations, based on optical character recognition (OCR), that support a single or two co-related document understanding tasks.
For example, RVL-CDIP \cite{harley2015icdar} contains annotations for DC, and FUNSD \cite{jaume2019funsd}
provides annotations for KEE and entity linking.
The majority of VRDU datasets, specifically those targeting forms and receipts, are designed for KEE.
\cite{davis-2019-deep,huang-2019-icdar,park2019cord,simsa2023docile,wang-2023-vrdu}.

\begin{table}[t!]
\centering
    \begin{tabular}{lllrrc}
        \bf Dataset & \bf Types & \bf Tasks & \bf Docs & \bf Labels & \bf OCR \\ \midrule
        CORD 
        & Receipts & $\mathcal{K}$ & $1,\!000$  & $30$ & \cmark \\
        DeepForm 
        & Receipts & $\mathcal{K}$ & $1,\!100$ & $5$ & \cmark \\
        DocILE\footnotemark 
        & Receipts & $\mathcal{K}$ & $7,\!000$ & $55$ & \cmark\\
        DocVQA 
        & Varied & $\mathcal{Q}$ & $12,\!767$ & $-$ & \cmark \\
        DUDE 
        & Varied & $\mathcal{Q}$ & $4,\!973$ & $-$ & \cmark \\
        FUNSD 
        & Forms & $\mathcal{K},\mathcal{L}$ & $199$ & $4$ & \cmark \\
        Kleister Char.
        & Reports & $\mathcal{K}$ & $540$ & $8$ & \cmark \\
        Klesiter NDA 
        & Legal & $\mathcal{K}$ & $2,\!778$ & $4$ & \cmark \\
        NAF 
        & Forms & $\mathcal{K},\mathcal{L}$  & $860$ & $14$ & \cmark \\
        RVL-CDIP 
        & Varied & $\mathcal{C}$ & $400,\!000$ & $16$ & \xmark \\
        SROIE 
        & Receipts & $\mathcal{K}$ & $1,\!000$ & $4$ & \cmark \\
        VRDU Ad-buy 
        & Receipts & $\mathcal{K}$ & $641$ & $10$ & \cmark \\
        VRDU Reg. 
        & Forms & $\mathcal{K}$ & $1,\!915$ & $6$ & \cmark \\
        \midrule
        \datasetName& Varied & $\mathcal{C},\mathcal{K}, \mathcal{Q}$ & $1,\!665$ & $69$ & \cmark
    \end{tabular}
    \caption{Existing VRDU dataset information. Tasks Legend: Document classification ($\mathcal{C}$), Entity linking ($\mathcal{L}$), Key entity extraction ($\mathcal{K}$), Visual question answering ($\mathcal{Q}$). The number of annotations is the total number of entities, links, questions, etc. provided by the dataset. Note that OCR is not available for the original versions of DeepForm, Kleister Charity, and Kleister NDA. However, \cite{Borchmann2021DUEED} provides OCR for these datasets.}
    \label{tab:datasets}
\end{table}

In this paper, we introduce \defn{\datasetName}, a new dataset comprised of $1,\!665$ publicly available structured business documents from US state government websites.
Our dataset is unique in that it tackles multiple distinct VRDU tasks: DC, KEE, and VQA.
Such a dataset is particularly beneficial to assess document processing in the wild, where requirements may necessitate models to perform several tasks on the same documents.
We created a hierarchical ontology of $69$ key entity classes over seven super categories that can be augmented with even more entity types in the future.
These provide a semantically rich and annotation-dense KEE dataset which enables us to construct a varied VQA set.
While similarly sized or larger VRDU datasets exist \cite{stanislawek-2021-kleister,simsa2023docile,wang-2023-vrdu}, they tend to focus on a single sort of document (e.g., receipts, NDA documents, etc.).
This may be insufficient for general purpose models that may be required in industry to accurately infer on a plethora of document types.
Therefore, \datasetName contributes a new large and varied dataset to the field.

\footnotetext{DocILE includes a further $100,\!000$ synthetic documents.}

\begin{figure*}[th!]
    \centering
    \includegraphics[width=4.3cm]{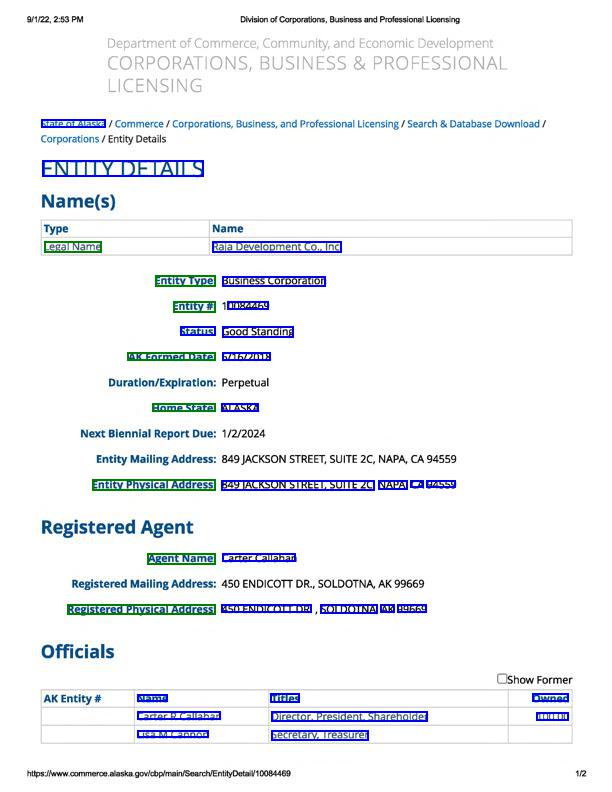}
    \includegraphics[width=4.3cm]{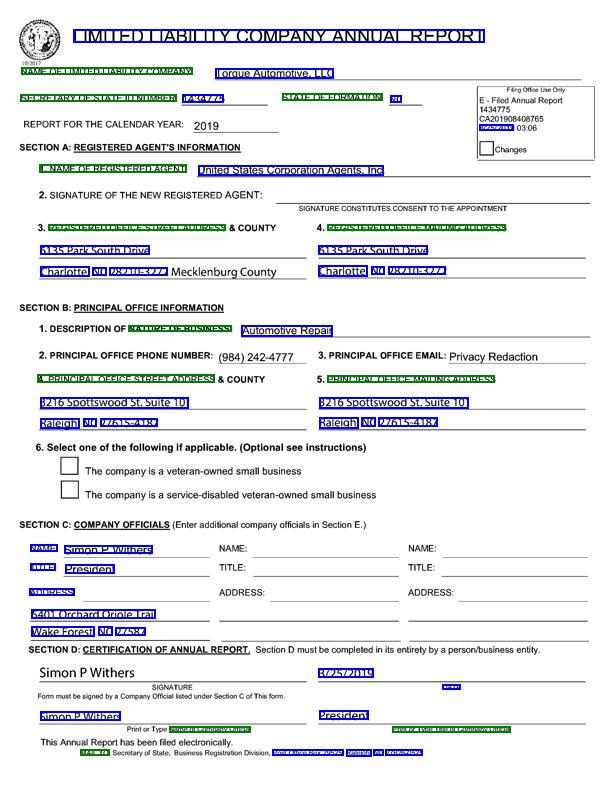}
    \includegraphics[width=4.3cm]{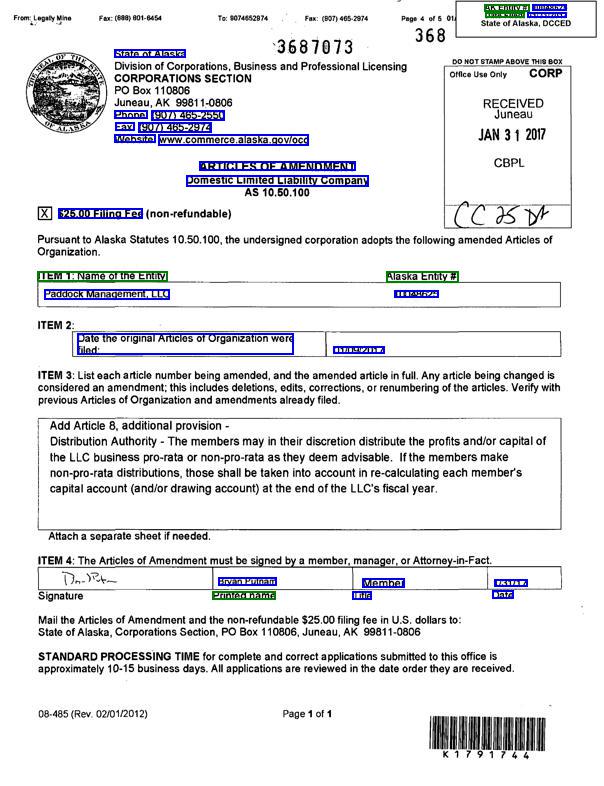}
    \includegraphics[width=4.3cm]{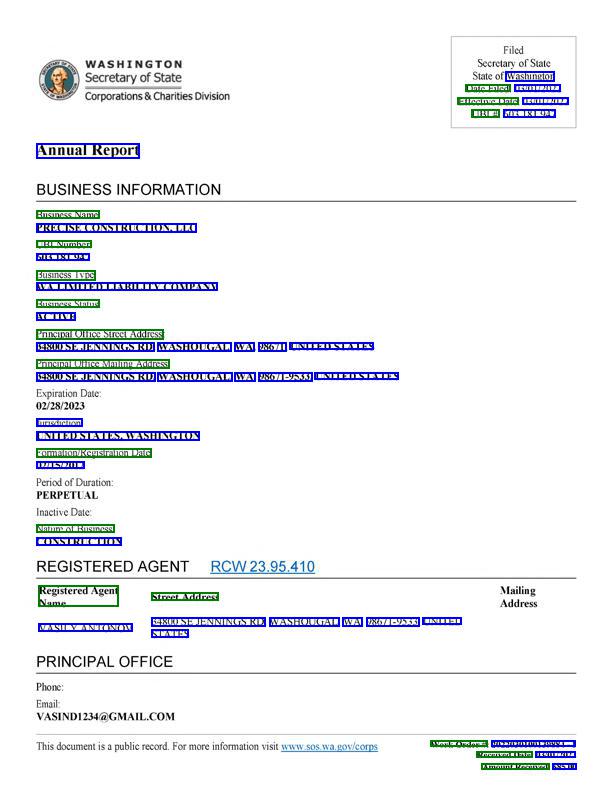}
    \caption{Examples of varied document styles in \datasetName with KEE annotations (entity labels are omitted for document format clarity).}
    \label{fig:doc-examples}
\end{figure*}

Our contributions are summarized below:
\begin{itemize}[noitemsep]
    \item We present \datasetName, a new annotated dataset consisting of $1,\!665$ structured business documents.
    \datasetName is the first VRDU dataset that supports three distinct tasks: DC, KEE, and VQA.
    Furthermore, it can be extended to facilitate multi-turn VQA, instruction tuning, and other downstream tasks with minimal additional effort. \datasetName is publicly available for non-commercial use.
    \item We provide six baselines for each task in \datasetName: two traditional text-only language models, BERT \cite{devlin-2019-bert} and RoBERTa \cite{liu-2020-roberta}. Two multi-modal language models, LayoutLM \cite{xu2020layoutlm} and LayoutLMv3 \cite{huang2022layoutlmv3}.
    And finally, two large language models (LLMs), GPT4 and DocLLM \cite{wang2023docllm}, where DocLLM incorporates multi-modal information into the language model.
    The best baseline across all tasks, DocLLM, achieves a DC F1 of $99.15$,
    KEE F1 score of $89.97$,
    and VQA ANLS score of $89.58$.
\end{itemize}

\section{Related Work}\label{sec:related}
In this section, we describe past datasets from the VRDU community as well as \sota models.

\subsection{Datasets}\label{sec:related:datasets}
The RVL-CDIP dataset \cite{harley2015icdar}, which consists of $400,\!000$ business related documents annotated for DC, is one of the first VRDU datasets that was released.
It solves an important but somewhat coarse-grained task, and RVL-CDIP is now mainly used to pre-train models.
Most modern VRDU datasets target information and entity extractions, which
were first introduced in 2019 when FUNSD \cite{jaume2019funsd}, SROIE \cite{huang-2019-icdar}, and CORD \cite{park2019cord} were released.
While the latter two focused on receipt documents, FUNSD
(\textbf{F}orm \textbf{U}nderstanding in \textbf{N}oisy \textbf{S}canned \textbf{D}ocuments) introduced the tasks of entity extraction and entity linking over forms.
It provided annotations of $199$ form documents from the RVL-CDIP dataset.
FUNSD annotates entities as \textit{question}, \textit{answer}, \textit{header}, or \textit{other}.
FUNSD is, however, more targeted at form structure extraction as its entities have structural rather than semantic meaning and are connected via entity linking.
FUNSD later received a revision that corrected annotation errors found in the original version \cite{vu-2020-revising}.\footnote{Furthermore, FUNSD has also been adapted for the task of form parsing \cite{zmigrod-2024-treeform}.}
While FUNSD is commonly used for VRDU fine tuning and evaluation,
its small size means it may be unreliable for comparing larger models \cite{Borchmann2021DUEED}.

CORD (\textbf{Co}nsolidated \textbf{R}eceipt \textbf{D}ataset) and SROIE (\textbf{S}canned \textbf{R}eceipt \textbf{O}CR and \textbf{I}nformation \textbf{E}xtraction) are KEE datasets for receipts.
SROIE contains $1,\!000$ documents with four semantic key entity labels that are commonly found in receipts.\footnote{SROIE also has text localisation and OCR annotations.}
CORD provides a richer key entity label set.
It consists of $1,\!000$ receipt documents that contain $30$ unique key entities subsumed by four super categories.\footnote{CORD claims to have $11,\!000$ documents, however only $1,\!000$ are publicly available.
Furthermore, there were originally $54$ unique entity labels over eight super categories \cite{park2019cord}. Some labels and super categories have been removed since the publication.}
Inspired by the CORD label ontology, we designed our own key entity label ontology in \cref{sec:dataset:entity}.
More recently, DocILE \cite{simsa2023docile}, a large dataset containing $7,\!000$ real-world receipts and $100,\!000$ synthetically generated receipts, annotated for KEE, has been introduced and used in the literature.
Their KEE task contains $55$ fine-grained labels.
Other KEE datasets cover additional document styles such as registration forms, NDAs, advertisements, \emph{inter alia} \cite{stanislawek-2021-kleister, wang-2023-vrdu}.

DocVQA \cite{mathew2021docvqa} introduced the task of VQA to the VRDU community.
The dataset is comprised of $12,\!767$ document images ($6,\!071$ total documents) from a wide variety of document types (e.g., forms, letters, and reports) with a total of $50,\!000$ questions.
Recently, a new document VQA dataset, DUDE \cite{landeghem-23-document}, has been proposed to offer a more varied VQA dataset.
We provide a detailed comparison of the datasets described above with our datatset, \datasetName, in \cref{tab:datasets}.\footnote{We discuss English datasets in this work. However, non-English VRDU datasets also exist \cite{qi-2022-dureadervis,xu-etal-2022-xfund}.
}

\begin{table*}[t]
    \centering
    \begin{tabular}{llrrrr} 
        \bf Class & \bf Examples & \bf Total & \bf Train & \bf Val & Test \\ \midrule
        Amendment Document & Article of Amend., Change of Address, Statement of Change & $85$ & $60$ & $9$ & $16$ \\ \midrule
        Application or Article & Application for Corporation, Article of Org., Name Reservation & $153$ & $111$ & $12$ & $30$ \\ \midrule
        Business Entity Details & Business Search Results, State Registry & $815$ & $570$ & $81$ & $164$ \\ \midrule
        Certificate or Statement & Certificate of Reinstatement, Statement of Good Standing & $90$ & $64$ & $9$ & $17$ \\ \midrule
        Periodical Report & Annual Report, Biennial Report & $522$ & $367$ & $50$ & $105$ \\ \midrule
        \bf Total & & $1,\!665$ & $1,\!172$ & $161$ & $332$
    \end{tabular}
    \caption{Example document titles and number of occurrences for \datasetName document classes.}
    \label{tab:doc-classes}
\end{table*}

\subsection{Models}\label{sec:related:models}
Early VRDU models incorporated textual and visual features in parallel and then merged them together.
Most commonly, a pre-trained transformer was used to embed spatially localized text and a pre-trained CNN-based model was used to encode the visual features \cite{denk2019bertgrid,wang2020docstruct,xu2020layoutlm,garncarek-21-lambert,lin2021vibertgrid,zhang2021trie}.
Subsequent models enabled richer interactions between text, spatial, and visual features by using a single multi-modal Transformer \cite{appalaraju2021docformer,powalski2021going,xu2021layoutlmv2,peng2022ernielayout,huang2022layoutlmv3,tang2023unifying}.

Other VRDU model architectures also exist in the literature.
For example, \cite{davis2021visual,zhang2022multimodal,lee-etal-2023-formnetv2} opted for graph-based approaches.
While graph-based methods still use the full multi-modal pipeline, some works have also discarded certain elements.
\cite{li-2021-structurallm,hong2022bros} abandoned visual features and instead solely relied on text and bounding box information.
On the other hand, a few recent models have experimented with vision only approaches to reduce the need of OCR \cite{davis-2022-end,kim-2022-ocr}.
Recently, LLMs have been increasingly used for VRDU tasks.
LLM architectures such as DocLLM \cite{wang2023docllm} make use of text and layout features, while models such as mPLUG-DocOwl \cite{ye2023mplugdocowl} leverage both text and general vision.

\section{The Business Document Dataset for Information Extraction}\label{sec:dataset}
In this paper, we introduce a new dataset for VRDU, \defn{\datasetName}, which consists of $1,\!665$ publicly available business documents.
In particular, we searched documents from US state websites (or their department of business website) which were under one of five document classes of interest shown in \cref{tab:doc-classes}. 
We obtained documents for Puerto Rico and all but eight of the $50$ states.\footnote{Documents from Illinois, Indiana, Louisiana, Maine, Mississippi, and Texas were blocked by a paywall and are thus not included in our dataset. Documents from Colorado and Michigan were available, but not for the distribution purposes of our work.}
The documents of \datasetName are 
partially structured, i.e., documents fall into styles such as forms, certificates, etc.
Examples of the varied structures and formats in the dataset are given in \cref{fig:doc-examples}.
\datasetName targets three prominent tasks in VRDU: DC, KEE, and VQA.
To the best of our knowledge, no current VRDU dataset tackles all three of these tasks, and no dataset of this size exists for KEE over multiple document types.
Furthermore, due to the rich and multi-task annotation scheme, our dataset has the potential to be extended in the future to support multi-turn VQA, instruction tuning, as well as other downstream VRDU tasks (e.g., entity linking), with minimal additional effort.
This could be of particular interest when considering multi-modal LLMs \cite{ye2023mplugdocowl, wang2023docllm}.
In the remainder of this section, we describe the data collection, annotation, and processing steps for each of the three tasks. Our annotation instructions are provided in \cref{app:annot}.

\paragraph{Availability and Usage of the Dataset}
\datasetName is publicly available for researchers to use for non-commercial purposes.\footnote{In order to get access to the dataset, please contact \href{mailto:airsyntheticdata.requests@jpmorgan.com}{airsyntheticdata.requests@jpmorgan.com}.}
Specific use policy can further be found in the licenses of the original data sources.\footnote{Links to these are provided in the dataset.}

\subsection{Document Processing}\label{sec:dataset:proc}
The initial collection for raw data yielded $1,\!890$ documents.
Many documents contained multiple pages, however, we only used the first page of each document in order to reduce annotation cost.\footnote{In many US state filings, the first page is the most complex in terms of layout and style.}
We used OCR to extract the text elements of each document.\footnote{Our annotation tool uses PDFPlumber to extract the OCR tokens and decide on a reading order. PDFPlumber is available at \url{https://github.com/jsvine/pdfplumber}.}
Throughout the annotation process described in \cref{sec:dataset:doc-class} and \cref{sec:dataset:entity}, $150$ documents were discarded due to poor OCR quality, lack of entities (fewer than five), or incompatibility with the document classes defined in \cref{tab:doc-classes}.
A further $75$ documents were discarded due to copyright issues.
After the annotation process, we created a train, validation, and test split of $70\%$, $20\%$, and $10\%$ respectively.
The split was done using stratified sampling on the document classes.\footnote{We note that in future iterations of the dataset, we also plan to release a train, validation, test split based on states, i.e., some states will be held out for the validation and test sets. This will work more towards assessing a model's ability to generalise to unseen document styles.}
Statistics regarding the entity types and document states are given in \cref{app:data}.

\begin{table*}[t]
    \centering
    \begin{tabular}{l l l r r r r}
        \bf Super Category & \bf Label & \bf Fine-grained Entity Examples & \bf Total & \bf Train & \bf Val & \bf Test \\ \midrule
        Business Entity & \texttt{ENT} & \texttt{ENT\_name}, \texttt{ENT\_number}, \texttt{ENT\_type} & $13,\!884$ & $9,\!703$ & $1,\!339$ & $2,\!842$ \\
        Entity Key Personnel & \texttt{KP} & \texttt{KP\_address\_street}, \texttt{KP\_name}, \texttt{KP\_title} & $9,\!845$ & $6,\!853$ & $906$ & $2,\!086$ \\
        File Attribute & \texttt{FILE} & \texttt{FILE\_date}, \texttt{FILE\_name}, \texttt{FILE\_number},  & $4,\!028$ & $2,\!840$ & $410$ & $778$ \\
        Government Official & \texttt{GO} & \texttt{GO\_adress\_city}, \texttt{GO\_name}, \texttt{GO\_title} & $3,\!046$ & $2,\!197$ & $280$ & $569$ \\
        Other & \texttt{OTHER} & \texttt{OTHER\_address}, \texttt{OTHER\_date}, \texttt{OTHER\_unknown}&  $839$ & $638$ & $48$ & $153$ \\
        Registered Agent & \texttt{AGT} & \texttt{AGT\_address\_city}, \texttt{AGT\_address\_state}, \texttt{AGT\_name} & $6,\!072$ & $4,\!248$ & $582$ & $1,\!242$ \\
        Signature & \texttt{SIG} & \texttt{SIG\_KP\_date}, \texttt{SIG\_KP\_printed\_name}, \texttt{SIG\_KP\_title} & $1,\!192$ & $850$ & $93$ & $249$ \\ \midrule
        \bf Total & & & $38,\!906$ & $27,\!329$ & $3,\!658$ & $7,\!919$
    \end{tabular}
    \caption{\datasetName key entity extraction super categories. For each super category, we provide the three most common fine-grained entity labels and the total number of occurrences of the super category.}
    \label{tab:entity-stats}
\end{table*}

\subsection{Document Classification}\label{sec:dataset:doc-class}
Document classification is the task of assigning a label to a document to denote its semantic or structural content. 
For example, RVL-CDIP categorises documents based on their style (e.g., form, letter, resume).
In BuDDIE, document classes have a semantic meaning.
The classes defined in \cref{tab:doc-classes} contain an underlying structural separation as well as semantic differences.
For instance, \emph{Business Entity Details} and \emph{Periodical Report} documents typically present a form-based format, while \emph{Certificate or Statement} documents tend to be more closely linked to letters.
There may exist semantic ambiguity and overlap between our classes; for example, an \emph{Article of Amendment} could be classified as \emph{Amendment Document} or \emph{Article or Application}.
Therefore, we constructed a list of ordered annotation rules for annotators to follow; we provide these rules in \cref{app:annot}.
In our above \emph{Article of Amendment} example, we rank the amendment documents higher than the other article documents, and so the document considered would fall into the \emph{Amendment Document} category.
We provide examples for each document class in \cref{tab:doc-classes}.

The DC annotation task was split between five annotators.
There were two rounds of annotation: (1) an initial annotation task to assign each document a class, and (2) a validation task to verify the labels that resulted from the initial round.
If there were repeated disagreements between an annotator and a validator, a third annotator would discuss discrepancies with both and decide on the final label based on the rules and discussions.
Documents for which no agreement was reached or which did not fall into any of our five document classes were discarded from the dataset.
In total, four documents were discarded due to the above reasons.

\subsection{Key Entity Extraction}\label{sec:dataset:entity}
Key entity extraction is the most popular task in VRDU.
The task is akin to a named entity recognition problem where each entity represents a key piece of information.
As documents vary in their content, KEE label sets tend to be large.
For example, CORD and DocILE, 
two 
similar datasets to ours, have label sets of $30$ and $55$ labels, respectively.
We offer a larger set of $69$ labels, since we focus on a wider domain (general business rather than receipts).
Like CORD and DocILE, we create our label set using super categories and specific detailed types.
In total, we consider six super categories: \textit{Business entity}, \textit{entity key personnel}, \textit{file attribute}, \textit{government official}, \textit{registered agent}, and \textit{signature}.
We additionally have an \textit{other} super category.
Under these seven super categories, we then have $69$ fine-grained labels.\footnote{We began with $74$ labels, and later discarded five fine-grained labels that had fewer than ten occurrences.}
We give frequency statistics for each of the super categories in \cref{tab:entity-stats} and a fine-grained analysis in \cref{app:data}.

The KEE annotation task was performed similarly to the DC annotation task.
The collection of documents was split between $12$ annotators who used
the PAWLS annotation tool \cite{neumann-etal-2021-pawls} to draw bounding boxes around relevant key entities.
Any OCR token that laid in the bounding box was then highlighted for the annotation.\footnote{The annotation tool also enabled free-form bounding boxes that were not bound to OCR tokens. While annotators were allowed to make such annotations, they were not included in this version of the dataset as our models assume the existence of OCR tokens.
A future version of this dataset may include free-form bounding boxes as well as OCR based bounding boxes.}
After the initial annotation round, each document was then validated by a different annotator.
If a validator found repeated inconsistencies with any annotations with regards to the annotation instructions, a third annotator would be consulted.
Any annotation in question either reached agreement across the three annotators or was discarded.
Annotators were instructed to only annotate an entity if they were confident in the specific annotation.
Consequently, our dataset may contain incomplete annotations as we put a stronger preference on the precision of our annotations.
We do not anticipate this to greatly impact the quality of our dataset given the high agreement score for KEE we describe in \cref{sec:dataset:quality}.

During this annotation task, annotators were also asked to mark entities that related to \textit{questions} or \textit{keys} in a form with a special \texttt{is\_key} tag.
We do not include these annotations in this version of \datasetName; however, we hope to include them in a future version, which will also feature entity linking and relation extraction as an additional task.
This would align \datasetName more closely to datasets such as FUNSD and NAF.

\begin{table}[t]
    \centering
    \begin{tabular}{lrrrr}
\bf Question Type & \bf Total & \bf Train & \bf Val & \bf Test \\
\midrule
Boolean No & $1,\!032$ & $739$ & $100$ & $193$ \\ 
Boolean Yes & $1,\!067$ & $742$ & $116$ & $209$ \\
Span & $6,\!571$ & $4,\!580$ & $674$ & $1,\!317$ \\
\midrule
\bf Total & $8,\!670$ & $6,\!061$ & $890$ & $1,\!719$
\end{tabular}
    \caption{Number of occurrences in the train, validation, and test splits of \datasetName for each type of question for VQA.}
    \label{tab:qa-stats}
\end{table}

\subsection{Visual Question Answering}\label{sec:dataset:vqa}
Question answering is a common NLP task where a model must provide a natural language response to a question given a passage \cite{yang-etal-2015-wikiqa,rajpurkar-etal-2016-squad,joshi-etal-2017-triviaqa,yang-etal-2018-hotpotqa}.
This naturally extends to images and evolves into VQA \cite{antol-2015-vqa}.
Document VQA is a mixture of these two tasks in which questions require understanding of both the text and visual properties of the document \cite{mathew2021docvqa}.

We consider two types of questions in \datasetName.
Firstly, \defn{span} questions are phrased as ``What is the \emph{X}?'', where \emph{X} is a key entity and the actual entity is the answer.
Secondly, \defn{boolean} questions are phrased as ``Is the \emph{X} \emph{Y}?'', where \emph{X} is a key entity as before and \emph{Y} is a candidate answer. These questions 
have \defn{yes} or \defn{no} answers.
Each key entity has an associated phrase to use in the question templates.
For example, questions for the the entity \texttt{AGT\_address\_zipcode} are phrased as ``What is the zip code of the registered agent?'' (for span questions) and ``Is the zip code of the registered agent 12345?'' (for boolean questions).

\begin{table*}[t]
\centering
\begin{tabular}{l c c c c c c c c}
    \multirow{2}{*}{\bf Model} & \multirow{2}{*}{\shortstack[c]{\bf Model \\ \\ \bf Size}} & \multicolumn{1}{c}{\bf Doc. Class.} & \multicolumn{3}{c}{\bf Key Entity Extraction} & \multicolumn{3}{c}{\bf Visual Question Ans.} \\ 
    & & \multicolumn{1}{c}{\bf F1 $\uparrow$}  & \multicolumn{1}{c}{\bf Prec. $\uparrow$} & \multicolumn{1}{c}{\bf Rec. $\uparrow$} & \multicolumn{1}{c}{\bf F1 $\uparrow$} & \multicolumn{1}{c}{\bf Acc. $\uparrow$} & \multicolumn{1}{c}{\bf ANLS $\uparrow$} & \multicolumn{1}{c}{\bf F1 $\uparrow$} \\
    \midrule
    BERT$_{\mathrm{base}}$ & $110$ M & $94.43$ & $80.94$ & $85.85$ & $83.32$ & $83.49$ & $86.54$ & $75.52$ \\
    RoBERTa$_{\mathrm{base}}$ & $125$ M & $91.96$ & $84.49$ & $87.48$ & $85.96$ & $84.28$ & $85.64$ & $90.06$ \\ \midrule
    LayoutLM$_{\mathrm{base}}$ & $160$ M & $96.01$ & $83.62$ & $88.16$ & $85.83$ & $54.95$ & $86.52$ & $75.32$ \\
    LayoutLMv3$_{\mathrm{base}}$ & $133$ M & $88.48$ & $84.23$ & $88.86$ & $86.49$ & $84.90$ & $86.85$ & $89.32$ \\ \midrule
    GPT4 & 
    --
    & $83.54$ & $77.76$ & $80.36$ & $77.76$ & $63.83$ & $80.05$ & $75.42$ \\
    DocLLM & $7$ B & $\boldsymbol{99.15}$ & $\textbf{90.55}$ & $\textbf{89.97}$ & $\textbf{89.97}$ & $\textbf{92.45}$ & $\textbf{89.58}$ & $\textbf{93.79}$ \\ 
    \end{tabular}
\caption{Baseline results on DC, KEE, and VQA for \datasetName. VQA accuracy considers Boolean questions while ANLS and F1 consider span questions. Note that GPT4 was run in a zero-shot context and DocLLM was instruction tuned using \datasetName along with other VRDU datasets. 
}
\label{tab:results}
\end{table*}

For each key entity observed in a document, we generate a question with a $30\%$ likelihood.
For the questions generated, $70\%$ are span questions and $30\%$ are boolean questions.
Span questions are generated by inserting the key entity phrase into the question template.
The answer is given as a list of key entity annotations\footnote{Each annotation corresponds to a set of OCR tokens.} as it is possible to observe multiple key entities of the same type in a document.\footnote{Past question answering datasets have allowed multiple spans to be a valid answer \cite{yang-etal-2018-hotpotqa}.}
We create a ``Yes'' question or a ``No'' question with equal probability.
In the case of a ``Yes'' question, the candidate answer is any of the annotations in the document with the specified entity label.\footnote{In this version of the dataset, we only generate boolean questions for key entities that only occur once in the document}
In the case of a ``No'' answer, we derive a candidate list from two sources.
Firstly, we consider other entities from the \emph{entire} dataset with the same fine-grained label (but not the same value).
Secondly, we consider key entities within the document that have the same key entity detailed type but not the same super category.
The candidate answer is chosen randomly from these two pools.
The total number of occurrences of each question type in the dataset is given in \cref{tab:qa-stats}.

\subsection{Annotation Quality}\label{sec:dataset:quality}
Using a sample of $60$ documents from \datasetName, we measure the agreement between the annotators and validators on each annotation task (DC and KEE). Following previous studies \cite{artstein-poesio-2008-survey, jochim-etal-2018-slide}, we sampled from a wide variety of annotations to mitigate some of the bias that could be caused by the sample size. We observe a Cohen's $\kappa$ of $0.976$ for document classification and $0.889$ for key entity extraction. Note that since the annotation validation was performed as a post-processing step to obtain the final \datasetName annotations, the agreement was computed using a sample of already-validated final annotations.\footnote{Importantly, the quality validators of the 60 sampled documents had not previously seen the documents in their original annotation or validation tasks.}
While our calculations may consequently provide an upper bound on Cohen's $\kappa$ for the original non-validated annotations, they yield a representative estimate of the quality of our final annotations.

\section{Experiments}\label{sec:experiments}
In this section we present baseline results for \datasetName.

\subsection{Baseline Models}\label{sec:experiments:baseline}
We consider six baseline models for our tasks.
BERT \cite{devlin-2019-bert} and RoBERTa \cite{liu-2020-roberta} are text-only models that solely rely on the OCR token sequence.
LayoutLM \cite{xu2020layoutlm} integrates additional spatial features into the transformer, and merges the transformer output with a vision CNN.
LayoutLMv3 \cite{huang2022layoutlmv3} incorporates vision features into the transformer architecture for each token.
For the aforementioned baselines, we finetune the base version of the model on each of the three tasks individually.\footnote{
Finetuning was done using the default hyperparameters of each respective model; a base learning rate of $10^{-4}$ was used with the Adam optimizer \cite{kingma-2014-adam}, and a batch size of four was selected.
All experiments were run with up to eight NVIDIA T4 GPUs. Smaller models used fewer GPUs.
}
In addition to the previous traditional baseline models, we further include two LLM baselines: GPT4 and DocLLM \cite{wang2023docllm}.
GPT4 is the text-only variant of the OpenAI model, to which we feed a document's OCR along with a prompt to represent the task at hand -- following the templates used in \cite{wang2023docllm}.
Lastly, DocLLM (based on Llama2 \cite{touvron2023llama}) is given the document's OCR along with spatial bounding box information and the task prompt.
GPT4 is used in a zero-shot setting while DocLLM has been instruction-tuned on the training split of \datasetName as well as other VRDU datasets.\footnote{Full detail regarding the training setup of DocLLM is described in the original manuscript \cite{wang2023docllm}.}
Due to cost and API usage constraints, we do not benchmark GPT4V on \datasetName. In addition, the discrepancy between the OCR tokens on which our annotations rely and GPT4V's proprietary image processor could potentially skew the scores of KEE and VQA token-level metrics.

\subsection{Evaluation Metrics}\label{sec:experiments:metrics}
We assess model performance on the three VRDU tasks of \datasetName with different metrics.
As our document classes are imbalanced in the dataset (see \cref{tab:doc-classes}), we report a macro F1 score for DC.
In other words, we take the mean F1 score across the five document classes.
For KEE, we report the weighted average token-level recall, precision, and F1 scores.
We also measure VQA performance using several metrics.
We evaluate boolean question performance using accuracy, and span questions using the Average Normalized Levenshtein Similarity (ANLS) and F1 scores.
The ANLS metric is a character-level metric used in \cite{mathew2021docvqa} whereas the F1 score gives the traditional token-level score.
These two metrics are reported separately to capture different aspects of measurement and granularity.

\subsection{Results}\label{sec:experiments:results}
\cref{tab:results} reports the performance of our baselines 
on \datasetName.\footnote{The experiments included a further $75$ documents from Colorado and Michigan. These documents were omitted from the public version of \datasetName due to distribution licenses.}
We note that the performance reported for GPT4 and DocLLM slightly differ from those in \cite{wang2023docllm}.
This is because the manuscript used accuracy rather than F1 for DC, included additional prompts for KEE that do not enable a fair comparison with non-LLM models, and aggregated results for VQA between span and boolean questions, which we separate in this paper.

With regards to DC, we observe strong performance from all models.
This was expected as certain keywords can be highly characteristic of specific document categories.
Furthermore, the imbalanced class distribution may further inflate performance even though we use macro F1.
We plan to add more fine-grained document classes in future versions of \datasetName as well as more documents to help alleviate the class imbalance.
Nevertheless, we observe a near perfect F1 score from DocLLM, showing that with the progress of multi-modal LLMs and the use of instruction-tuning, the \datasetName DC task may be solved to human level performance. Indeed, the DocLLM F1 for DC is higher than the inter-annotator agreement for the same task.

For KEE, we observe that the spatially aware models (LayoutLM, LayoutLMv3, and DocLLM) tend to have a much better recall than their text-only counterparts.
While GPT4 demonstrates the worst result, the spatially aware LLM, DocLLM, outperforms any of the dedicated smaller models.
Note that GPT4's scores are still considerably resilient given the zero-shot setting, as opposed to the fine-tuned setting for the other models.

The VQA F1 scores exhibit high variability in the reported results.
This can be attributed to the inherent fluctuation in token-level evaluation compared to the character-based ANLS metric.
Specifically, we observe a large discrepancy between the VQA F1 scores of BERT, LayoutLM, and GPT4 with respect to the other models.
We hypothesize that the performance of the first two is due to a difference in tokenizers used.
Specifically, LayoutLM and BERT employ a word-piece tokenizer, whereas the other models employ a Byte-Pair Encoding (BPE) tokenizer.
The BPE tokenizer is likely to capture tokens with greater accuracy, consequently leading to improved F1 scores.
It is probable again that GPT4's relatively low performance across all VQA metrics can be attested to both its lack of input layout information and to the zero-shot inference (the model sometimes extracts less or more context than expected in the annotations).
DocLLM once again outperforms the other models on VQA, specifically in terms of the boolean question accuracy.

\section{Conclusion}\label{sec:conclusion}
In this paper, we introduced a new VRDU dataset, \datasetName, consisting of $1,\!665$ annotated documents.
\datasetName is unique in its varied document styles, sizes, and annotations for three distinct tasks.
We use a variety of language models, multi-modal language models, and LLMs to provide comprehensive baselines for our dataset.
While we note DocLLM's impressive performance across the tasks,
VRDU model performance is still
not comparable to human performance on the tasks of KEE and VQA \cite{mathew2021docvqa}, and zero-shot prompted LLMs still have room for improvement.
We hope that our dataset can be a valuable resource which can challenge the research community to seek more robust VRDU models, and will encourage further research in this domain.
Furthermore, we hope that future work on \datasetName will include multi-page annotations, multi-turn VQA, and instruction tuning.

\section*{Disclaimer}
This paper was prepared for informational purposes by the Artificial Intelligence Research group of JPMorgan Chase \& Co and its affiliates (“JP Morgan”), and is not a product of the Research Department of JP Morgan. JP Morgan makes no representation and warranty whatsoever and disclaims all liability, for the completeness, accuracy or reliability of the information contained herein. This document is not intended as investment research or investment advice, or a recommendation, offer or solicitation for the purchase or sale of any security, financial instrument, financial product or service, or to be used in any way for evaluating the merits of participating in any transaction, and shall not constitute a solicitation under any jurisdiction or to any person, if such solicitation under such jurisdiction or to such person would be unlawful.

\bibliographystyle{ACM-Reference-Format}
\bibliography{references.bib}


\begin{thebibliography}{48}


\ifx \showCODEN    \undefined \def \showCODEN     #1{\unskip}     \fi
\ifx \showDOI      \undefined \def \showDOI       #1{#1}\fi
\ifx \showISBNx    \undefined \def \showISBNx     #1{\unskip}     \fi
\ifx \showISBNxiii \undefined \def \showISBNxiii  #1{\unskip}     \fi
\ifx \showISSN     \undefined \def \showISSN      #1{\unskip}     \fi
\ifx \showLCCN     \undefined \def \showLCCN      #1{\unskip}     \fi
\ifx \shownote     \undefined \def \shownote      #1{#1}          \fi
\ifx \showarticletitle \undefined \def \showarticletitle #1{#1}   \fi
\ifx \showURL      \undefined \def \showURL       {\relax}        \fi
\providecommand\bibfield[2]{#2}
\providecommand\bibinfo[2]{#2}
\providecommand\natexlab[1]{#1}
\providecommand\showeprint[2][]{arXiv:#2}

\bibitem[Antol et~al\mbox{.}(2015)]%
        {antol-2015-vqa}
\bibfield{author}{\bibinfo{person}{Stanislaw Antol}, \bibinfo{person}{Aishwarya Agrawal}, \bibinfo{person}{Jiasen Lu}, \bibinfo{person}{Margaret Mitchell}, \bibinfo{person}{Dhruv Batra}, \bibinfo{person}{C.~Lawrence Zitnick}, {and} \bibinfo{person}{Devi Parikh}.} \bibinfo{year}{2015}\natexlab{}.
\newblock \showarticletitle{{VQA:} {V}isual Question Answering}. In \bibinfo{booktitle}{\emph{2015 {IEEE} International Conference on Computer Vision, {ICCV} 2015, Santiago, Chile, December 7-13, 2015}}. \bibinfo{publisher}{{IEEE} Computer Society}, \bibinfo{pages}{2425--2433}.
\newblock
\urldef\tempurl%
\url{https://doi.org/10.1109/ICCV.2015.279}
\showDOI{\tempurl}


\bibitem[Appalaraju et~al\mbox{.}(2021)]%
        {appalaraju2021docformer}
\bibfield{author}{\bibinfo{person}{Srikar Appalaraju}, \bibinfo{person}{Bhavan Jasani}, \bibinfo{person}{Bhargava~Urala Kota}, \bibinfo{person}{Yusheng Xie}, {and} \bibinfo{person}{R. Manmatha}.} \bibinfo{year}{2021}\natexlab{}.
\newblock \showarticletitle{{D}oc{F}ormer: End-to-End Transformer for Document Understanding}. In \bibinfo{booktitle}{\emph{2021 {IEEE/CVF} International Conference on Computer Vision, {ICCV} 2021, Montreal, QC, Canada, October 10-17, 2021}}. \bibinfo{publisher}{{IEEE}}, \bibinfo{pages}{973--983}.
\newblock
\urldef\tempurl%
\url{https://doi.org/10.1109/ICCV48922.2021.00103}
\showDOI{\tempurl}


\bibitem[Artstein and Poesio(2008)]%
        {artstein-poesio-2008-survey}
\bibfield{author}{\bibinfo{person}{Ron Artstein} {and} \bibinfo{person}{Massimo Poesio}.} \bibinfo{year}{2008}\natexlab{}.
\newblock \showarticletitle{Survey Article: Inter-Coder Agreement for Computational Linguistics}.
\newblock \bibinfo{journal}{\emph{Computational Linguistics}} \bibinfo{volume}{34}, \bibinfo{number}{4} (\bibinfo{year}{2008}), \bibinfo{pages}{555--596}.
\newblock
\urldef\tempurl%
\url{https://doi.org/10.1162/coli.07-034-R2}
\showDOI{\tempurl}


\bibitem[Borchmann et~al\mbox{.}(2021)]%
        {Borchmann2021DUEED}
\bibfield{author}{\bibinfo{person}{Lukasz Borchmann}, \bibinfo{person}{Michal Pietruszka}, \bibinfo{person}{Tomasz Stanislawek}, \bibinfo{person}{Dawid Jurkiewicz}, \bibinfo{person}{Michal Turski}, \bibinfo{person}{Karolina Szyndler}, {and} \bibinfo{person}{Filip Gralinski}.} \bibinfo{year}{2021}\natexlab{}.
\newblock \showarticletitle{{DUE:} End-to-End Document Understanding Benchmark}. In \bibinfo{booktitle}{\emph{Proceedings of the Neural Information Processing Systems Track on Datasets and Benchmarks 1, NeurIPS Datasets and Benchmarks 2021, December 2021, virtual}}, \bibfield{editor}{\bibinfo{person}{Joaquin Vanschoren} {and} \bibinfo{person}{Sai{-}Kit Yeung}} (Eds.).
\newblock
\urldef\tempurl%
\url{https://datasets-benchmarks-proceedings.neurips.cc/paper/2021/hash/069059b7ef840f0c74a814ec9237b6ec-Abstract-round2.html}
\showURL{%
\tempurl}


\bibitem[Davis et~al\mbox{.}(2019)]%
        {davis-2019-deep}
\bibfield{author}{\bibinfo{person}{Brian~L. Davis}, \bibinfo{person}{Bryan~S. Morse}, \bibinfo{person}{Scott Cohen}, \bibinfo{person}{Brian~L. Price}, {and} \bibinfo{person}{Chris Tensmeyer}.} \bibinfo{year}{2019}\natexlab{}.
\newblock \showarticletitle{Deep Visual Template-Free Form Parsing}. In \bibinfo{booktitle}{\emph{2019 International Conference on Document Analysis and Recognition, {ICDAR} 2019, Sydney, Australia, September 20-25, 2019}}. \bibinfo{publisher}{{IEEE}}, \bibinfo{pages}{134--141}.
\newblock
\urldef\tempurl%
\url{https://doi.org/10.1109/ICDAR.2019.00030}
\showDOI{\tempurl}


\bibitem[Davis et~al\mbox{.}(2021)]%
        {davis2021visual}
\bibfield{author}{\bibinfo{person}{Brian~L. Davis}, \bibinfo{person}{Bryan~S. Morse}, \bibinfo{person}{Brian~L. Price}, \bibinfo{person}{Chris Tensmeyer}, {and} \bibinfo{person}{Curtis Wigington}.} \bibinfo{year}{2021}\natexlab{}.
\newblock \showarticletitle{{V}isual {FUDGE}: {F}orm Understanding via Dynamic Graph Editing}. In \bibinfo{booktitle}{\emph{16th International Conference on Document Analysis and Recognition, {ICDAR} 2021, Lausanne, Switzerland, September 5-10, 2021, Proceedings, Part {I}}} \emph{(\bibinfo{series}{Lecture Notes in Computer Science}, Vol.~\bibinfo{volume}{12821})}, \bibfield{editor}{\bibinfo{person}{Josep Llad{\'{o}}s}, \bibinfo{person}{Daniel Lopresti}, {and} \bibinfo{person}{Seiichi Uchida}} (Eds.). \bibinfo{publisher}{Springer}, \bibinfo{pages}{416--431}.
\newblock
\urldef\tempurl%
\url{https://doi.org/10.1007/978-3-030-86549-8\_27}
\showDOI{\tempurl}


\bibitem[Davis et~al\mbox{.}(2022)]%
        {davis-2022-end}
\bibfield{author}{\bibinfo{person}{Brian~L. Davis}, \bibinfo{person}{Bryan~S. Morse}, \bibinfo{person}{Brian~L. Price}, \bibinfo{person}{Chris Tensmeyer}, \bibinfo{person}{Curtis Wigington}, {and} \bibinfo{person}{Vlad~I. Morariu}.} \bibinfo{year}{2022}\natexlab{}.
\newblock \showarticletitle{End-to-End Document Recognition and Understanding with Dessurt}. In \bibinfo{booktitle}{\emph{Computer Vision - {ECCV} 2022 Workshops - Tel Aviv, Israel, October 23-27, 2022, Proceedings, Part {IV}}} \emph{(\bibinfo{series}{Lecture Notes in Computer Science}, Vol.~\bibinfo{volume}{13804})}, \bibfield{editor}{\bibinfo{person}{Leonid Karlinsky}, \bibinfo{person}{Tomer Michaeli}, {and} \bibinfo{person}{Ko~Nishino}} (Eds.). \bibinfo{publisher}{Springer}, \bibinfo{pages}{280--296}.
\newblock
\urldef\tempurl%
\url{https://doi.org/10.1007/978-3-031-25069-9\_19}
\showDOI{\tempurl}


\bibitem[Denk and Reisswig(2019)]%
        {denk2019bertgrid}
\bibfield{author}{\bibinfo{person}{Timo~I. Denk} {and} \bibinfo{person}{Christian Reisswig}.} \bibinfo{year}{2019}\natexlab{}.
\newblock \showarticletitle{{BERT}grid: Contextualized Embedding for 2{D} Document Representation and Understanding}.
\newblock \bibinfo{journal}{\emph{CoRR}}  \bibinfo{volume}{abs/1909.04948} (\bibinfo{year}{2019}).
\newblock
\showeprint[arXiv]{1909.04948}
\urldef\tempurl%
\url{http://arxiv.org/abs/1909.04948}
\showURL{%
\tempurl}


\bibitem[Devlin et~al\mbox{.}(2019)]%
        {devlin-2019-bert}
\bibfield{author}{\bibinfo{person}{Jacob Devlin}, \bibinfo{person}{Ming{-}Wei Chang}, \bibinfo{person}{Kenton Lee}, {and} \bibinfo{person}{Kristina Toutanova}.} \bibinfo{year}{2019}\natexlab{}.
\newblock \showarticletitle{{BERT}: {P}re-training of Deep Bidirectional Transformers for Language Understanding}. In \bibinfo{booktitle}{\emph{Proceedings of the 2019 Conference of the North American Chapter of the Association for Computational Linguistics: Human Language Technologies, {NAACL-HLT} 2019, Minneapolis, MN, USA, June 2-7, 2019, Volume 1 (Long and Short Papers)}}, \bibfield{editor}{\bibinfo{person}{Jill Burstein}, \bibinfo{person}{Christy Doran}, {and} \bibinfo{person}{Thamar Solorio}} (Eds.). \bibinfo{publisher}{Association for Computational Linguistics}, \bibinfo{pages}{4171--4186}.
\newblock
\urldef\tempurl%
\url{https://doi.org/10.18653/v1/n19-1423}
\showDOI{\tempurl}


\bibitem[Garncarek et~al\mbox{.}(2021)]%
        {garncarek-21-lambert}
\bibfield{author}{\bibinfo{person}{{\L}ukasz Garncarek}, \bibinfo{person}{Rafa{\l} Powalski}, \bibinfo{person}{Tomasz Stanis{\l}awek}, \bibinfo{person}{Bartosz Topolski}, \bibinfo{person}{Piotr Halama}, \bibinfo{person}{Micha{\l} Turski}, {and} \bibinfo{person}{Filip Grali{\'{n}}ski}.} \bibinfo{year}{2021}\natexlab{}.
\newblock \showarticletitle{{LAMBERT}: {L}ayout-Aware Language Modeling for Information Extraction}. In \bibinfo{booktitle}{\emph{Document Analysis and Recognition -- ICDAR 2021}}, \bibfield{editor}{\bibinfo{person}{Josep Llad{\'o}s}, \bibinfo{person}{Daniel Lopresti}, {and} \bibinfo{person}{Seiichi Uchida}} (Eds.). \bibinfo{publisher}{Springer International Publishing}, \bibinfo{address}{Cham}, \bibinfo{pages}{532--547}.
\newblock
\showISBNx{978-3-030-86549-8}


\bibitem[Harley et~al\mbox{.}(2015)]%
        {harley2015icdar}
\bibfield{author}{\bibinfo{person}{Adam~W. Harley}, \bibinfo{person}{Alex Ufkes}, {and} \bibinfo{person}{Konstantinos~G. Derpanis}.} \bibinfo{year}{2015}\natexlab{}.
\newblock \showarticletitle{Evaluation of deep convolutional nets for document image classification and retrieval}. In \bibinfo{booktitle}{\emph{13th International Conference on Document Analysis and Recognition, {ICDAR} 2015, Nancy, France, August 23-26, 2015}}. \bibinfo{publisher}{{IEEE} Computer Society}, \bibinfo{pages}{991--995}.
\newblock
\urldef\tempurl%
\url{https://doi.org/10.1109/ICDAR.2015.7333910}
\showDOI{\tempurl}


\bibitem[Hong et~al\mbox{.}(2022)]%
        {hong2022bros}
\bibfield{author}{\bibinfo{person}{Teakgyu Hong}, \bibinfo{person}{Donghyun Kim}, \bibinfo{person}{Mingi Ji}, \bibinfo{person}{Wonseok Hwang}, \bibinfo{person}{Daehyun Nam}, {and} \bibinfo{person}{Sungrae Park}.} \bibinfo{year}{2022}\natexlab{}.
\newblock \showarticletitle{{BROS:} {A} Pre-trained Language Model Focusing on Text and Layout for Better Key Information Extraction from Documents}. In \bibinfo{booktitle}{\emph{Thirty-Sixth {AAAI} Conference on Artificial Intelligence, {AAAI} 2022, Thirty-Fourth Conference on Innovative Applications of Artificial Intelligence, {IAAI} 2022, The Twelveth Symposium on Educational Advances in Artificial Intelligence, {EAAI} 2022 Virtual Event, February 22 - March 1, 2022}}. \bibinfo{publisher}{{AAAI} Press}, \bibinfo{pages}{10767--10775}.
\newblock
\urldef\tempurl%
\url{https://ojs.aaai.org/index.php/AAAI/article/view/21322}
\showURL{%
\tempurl}


\bibitem[Huang et~al\mbox{.}(2022)]%
        {huang2022layoutlmv3}
\bibfield{author}{\bibinfo{person}{Yupan Huang}, \bibinfo{person}{Tengchao Lv}, \bibinfo{person}{Lei Cui}, \bibinfo{person}{Yutong Lu}, {and} \bibinfo{person}{Furu Wei}.} \bibinfo{year}{2022}\natexlab{}.
\newblock \showarticletitle{{L}ayout{LM}v3: Pre-training for Document {AI} with Unified Text and Image Masking}. In \bibinfo{booktitle}{\emph{{MM} '22: The 30th {ACM} International Conference on Multimedia, Lisboa, Portugal, October 10 - 14, 2022}}, \bibfield{editor}{\bibinfo{person}{Jo{\~{a}}o Magalh{\~{a}}es}, \bibinfo{person}{Alberto~Del Bimbo}, \bibinfo{person}{Shin'ichi Satoh}, \bibinfo{person}{Nicu Sebe}, \bibinfo{person}{Xavier Alameda{-}Pineda}, \bibinfo{person}{Qin Jin}, \bibinfo{person}{Vincent Oria}, {and} \bibinfo{person}{Laura Toni}} (Eds.). \bibinfo{publisher}{{ACM}}, \bibinfo{pages}{4083--4091}.
\newblock
\urldef\tempurl%
\url{https://doi.org/10.1145/3503161.3548112}
\showDOI{\tempurl}


\bibitem[Huang et~al\mbox{.}(2019)]%
        {huang-2019-icdar}
\bibfield{author}{\bibinfo{person}{Zheng Huang}, \bibinfo{person}{Kai Chen}, \bibinfo{person}{Jianhua He}, \bibinfo{person}{Xiang Bai}, \bibinfo{person}{Dimosthenis Karatzas}, \bibinfo{person}{Shijian Lu}, {and} \bibinfo{person}{C.~V. Jawahar}.} \bibinfo{year}{2019}\natexlab{}.
\newblock \showarticletitle{{ICDAR2019} Competition on Scanned Receipt {OCR} and Information Extraction}. In \bibinfo{booktitle}{\emph{2019 International Conference on Document Analysis and Recognition, {ICDAR} 2019, Sydney, Australia, September 20-25, 2019}}. \bibinfo{publisher}{{IEEE}}, \bibinfo{pages}{1516--1520}.
\newblock
\urldef\tempurl%
\url{https://doi.org/10.1109/ICDAR.2019.00244}
\showDOI{\tempurl}


\bibitem[Jaume et~al\mbox{.}(2019)]%
        {jaume2019funsd}
\bibfield{author}{\bibinfo{person}{Guillaume Jaume}, \bibinfo{person}{Hazim~Kemal Ekenel}, {and} \bibinfo{person}{Jean{-}Philippe Thiran}.} \bibinfo{year}{2019}\natexlab{}.
\newblock \showarticletitle{{FUNSD:} {A} Dataset for Form Understanding in Noisy Scanned Documents}. In \bibinfo{booktitle}{\emph{2nd International Workshop on Open Services and Tools for Document Analysis, OST@ICDAR 2019, Sydney, Australia, September 22-25, 2019}}. \bibinfo{publisher}{{IEEE}}, \bibinfo{pages}{1--6}.
\newblock
\urldef\tempurl%
\url{https://doi.org/10.1109/ICDARW.2019.10029}
\showDOI{\tempurl}


\bibitem[Jochim et~al\mbox{.}(2018)]%
        {jochim-etal-2018-slide}
\bibfield{author}{\bibinfo{person}{Charles Jochim}, \bibinfo{person}{Francesca Bonin}, \bibinfo{person}{Roy Bar-Haim}, {and} \bibinfo{person}{Noam Slonim}.} \bibinfo{year}{2018}\natexlab{}.
\newblock \showarticletitle{{SLIDE} - a Sentiment Lexicon of Common Idioms}. In \bibinfo{booktitle}{\emph{Proceedings of the Eleventh International Conference on Language Resources and Evaluation ({LREC} 2018)}}, \bibfield{editor}{\bibinfo{person}{Nicoletta Calzolari}, \bibinfo{person}{Khalid Choukri}, \bibinfo{person}{Christopher Cieri}, \bibinfo{person}{Thierry Declerck}, \bibinfo{person}{Sara Goggi}, \bibinfo{person}{Koiti Hasida}, \bibinfo{person}{Hitoshi Isahara}, \bibinfo{person}{Bente Maegaard}, \bibinfo{person}{Joseph Mariani}, \bibinfo{person}{H{\'e}l{\`e}ne Mazo}, \bibinfo{person}{Asuncion Moreno}, \bibinfo{person}{Jan Odijk}, \bibinfo{person}{Stelios Piperidis}, {and} \bibinfo{person}{Takenobu Tokunaga}} (Eds.). \bibinfo{publisher}{European Language Resources Association (ELRA)}, \bibinfo{address}{Miyazaki, Japan}.
\newblock
\urldef\tempurl%
\url{https://aclanthology.org/L18-1379}
\showURL{%
\tempurl}


\bibitem[Joshi et~al\mbox{.}(2017)]%
        {joshi-etal-2017-triviaqa}
\bibfield{author}{\bibinfo{person}{Mandar Joshi}, \bibinfo{person}{Eunsol Choi}, \bibinfo{person}{Daniel Weld}, {and} \bibinfo{person}{Luke Zettlemoyer}.} \bibinfo{year}{2017}\natexlab{}.
\newblock \showarticletitle{{T}rivia{QA}: A Large Scale Distantly Supervised Challenge Dataset for Reading Comprehension}. In \bibinfo{booktitle}{\emph{Proceedings of the 55th Annual Meeting of the Association for Computational Linguistics (Volume 1: Long Papers)}}. \bibinfo{publisher}{Association for Computational Linguistics}, \bibinfo{address}{Vancouver, Canada}, \bibinfo{pages}{1601--1611}.
\newblock
\urldef\tempurl%
\url{https://doi.org/10.18653/v1/P17-1147}
\showDOI{\tempurl}


\bibitem[Kim et~al\mbox{.}(2022)]%
        {kim-2022-ocr}
\bibfield{author}{\bibinfo{person}{Geewook Kim}, \bibinfo{person}{Teakgyu Hong}, \bibinfo{person}{Moonbin Yim}, \bibinfo{person}{JeongYeon Nam}, \bibinfo{person}{Jinyoung Park}, \bibinfo{person}{Jinyeong Yim}, \bibinfo{person}{Wonseok Hwang}, \bibinfo{person}{Sangdoo Yun}, \bibinfo{person}{Dongyoon Han}, {and} \bibinfo{person}{Seunghyun Park}.} \bibinfo{year}{2022}\natexlab{}.
\newblock \showarticletitle{{OCR}-Free Document Understanding Transformer}. In \bibinfo{booktitle}{\emph{Computer Vision - {ECCV} 2022 - 17th European Conference, Tel Aviv, Israel, October 23-27, 2022, Proceedings, Part {XXVIII}}} \emph{(\bibinfo{series}{Lecture Notes in Computer Science}, Vol.~\bibinfo{volume}{13688})}, \bibfield{editor}{\bibinfo{person}{Shai Avidan}, \bibinfo{person}{Gabriel~J. Brostow}, \bibinfo{person}{Moustapha Ciss{\'{e}}}, \bibinfo{person}{Giovanni~Maria Farinella}, {and} \bibinfo{person}{Tal Hassner}} (Eds.). \bibinfo{publisher}{Springer}, \bibinfo{pages}{498--517}.
\newblock
\urldef\tempurl%
\url{https://doi.org/10.1007/978-3-031-19815-1\_29}
\showDOI{\tempurl}


\bibitem[Kingma and Ba(2015)]%
        {kingma-2014-adam}
\bibfield{author}{\bibinfo{person}{Diederik~P. Kingma} {and} \bibinfo{person}{Jimmy Ba}.} \bibinfo{year}{2015}\natexlab{}.
\newblock \showarticletitle{Adam: {A} Method for Stochastic Optimization}. In \bibinfo{booktitle}{\emph{3rd International Conference on Learning Representations, {ICLR} 2015, San Diego, CA, USA, May 7-9, 2015, Conference Track Proceedings}}, \bibfield{editor}{\bibinfo{person}{Yoshua Bengio} {and} \bibinfo{person}{Yann LeCun}} (Eds.).
\newblock
\urldef\tempurl%
\url{http://arxiv.org/abs/1412.6980}
\showURL{%
\tempurl}


\bibitem[Landeghem et~al\mbox{.}(2023)]%
        {landeghem-23-document}
\bibfield{author}{\bibinfo{person}{Jordy~Van Landeghem}, \bibinfo{person}{Rub{\`{e}}n Tito}, \bibinfo{person}{Lukasz Borchmann}, \bibinfo{person}{Michal Pietruszka}, \bibinfo{person}{Pawel J{\'{o}}ziak}, \bibinfo{person}{Rafal Powalski}, \bibinfo{person}{Dawid Jurkiewicz}, \bibinfo{person}{Micka{\"{e}}l Coustaty}, \bibinfo{person}{Bertrand Anckaert}, \bibinfo{person}{Ernest Valveny}, \bibinfo{person}{Matthew~B. Blaschko}, \bibinfo{person}{Sien Moens}, {and} \bibinfo{person}{Tomasz Stanislawek}.} \bibinfo{year}{2023}\natexlab{}.
\newblock \showarticletitle{Document Understanding Dataset and Evaluation {(DUDE)}}.
\newblock \bibinfo{journal}{\emph{CoRR}}  \bibinfo{volume}{abs/2305.08455} (\bibinfo{year}{2023}).
\newblock
\urldef\tempurl%
\url{https://doi.org/10.48550/arXiv.2305.08455}
\showDOI{\tempurl}
\showeprint[arXiv]{2305.08455}


\bibitem[Lee et~al\mbox{.}(2023)]%
        {lee-etal-2023-formnetv2}
\bibfield{author}{\bibinfo{person}{Chen-Yu Lee}, \bibinfo{person}{Chun-Liang Li}, \bibinfo{person}{Hao Zhang}, \bibinfo{person}{Timothy Dozat}, \bibinfo{person}{Vincent Perot}, \bibinfo{person}{Guolong Su}, \bibinfo{person}{Xiang Zhang}, \bibinfo{person}{Kihyuk Sohn}, \bibinfo{person}{Nikolay Glushnev}, \bibinfo{person}{Renshen Wang}, \bibinfo{person}{Joshua Ainslie}, \bibinfo{person}{Shangbang Long}, \bibinfo{person}{Siyang Qin}, \bibinfo{person}{Yasuhisa Fujii}, \bibinfo{person}{Nan Hua}, {and} \bibinfo{person}{Tomas Pfister}.} \bibinfo{year}{2023}\natexlab{}.
\newblock \showarticletitle{{F}orm{N}et{V}2: Multimodal Graph Contrastive Learning for Form Document Information Extraction}. In \bibinfo{booktitle}{\emph{Proceedings of the 61st Annual Meeting of the Association for Computational Linguistics (Volume 1: Long Papers)}}. \bibinfo{publisher}{Association for Computational Linguistics}, \bibinfo{address}{Toronto, Canada}, \bibinfo{pages}{9011--9026}.
\newblock
\urldef\tempurl%
\url{https://doi.org/10.18653/v1/2023.acl-long.501}
\showDOI{\tempurl}


\bibitem[Li et~al\mbox{.}(2021)]%
        {li-2021-structurallm}
\bibfield{author}{\bibinfo{person}{Chenliang Li}, \bibinfo{person}{Bin Bi}, \bibinfo{person}{Ming Yan}, \bibinfo{person}{Wei Wang}, \bibinfo{person}{Songfang Huang}, \bibinfo{person}{Fei Huang}, {and} \bibinfo{person}{Luo Si}.} \bibinfo{year}{2021}\natexlab{}.
\newblock \showarticletitle{{S}tructural{LM}: Structural Pre-training for Form Understanding}. In \bibinfo{booktitle}{\emph{Proceedings of the 59th Annual Meeting of the Association for Computational Linguistics and the 11th International Joint Conference on Natural Language Processing (Volume 1: Long Papers)}}. \bibinfo{publisher}{Association for Computational Linguistics}, \bibinfo{address}{Online}, \bibinfo{pages}{6309--6318}.
\newblock
\urldef\tempurl%
\url{https://doi.org/10.18653/v1/2021.acl-long.493}
\showDOI{\tempurl}


\bibitem[Lin et~al\mbox{.}(2021)]%
        {lin2021vibertgrid}
\bibfield{author}{\bibinfo{person}{Weihong Lin}, \bibinfo{person}{Qifang Gao}, \bibinfo{person}{Lei Sun}, \bibinfo{person}{Zhuoyao Zhong}, \bibinfo{person}{Kai Hu}, \bibinfo{person}{Qin Ren}, {and} \bibinfo{person}{Qiang Huo}.} \bibinfo{year}{2021}\natexlab{}.
\newblock \showarticletitle{{V}i{BERT}grid: {A} Jointly Trained Multi-modal 2{D} Document Representation for Key Information Extraction from Documents}. In \bibinfo{booktitle}{\emph{16th International Conference on Document Analysis and Recognition, {ICDAR} 2021, Lausanne, Switzerland, September 5-10, 2021, Proceedings, Part {I}}} \emph{(\bibinfo{series}{Lecture Notes in Computer Science}, Vol.~\bibinfo{volume}{12821})}, \bibfield{editor}{\bibinfo{person}{Josep Llad{\'{o}}s}, \bibinfo{person}{Daniel Lopresti}, {and} \bibinfo{person}{Seiichi Uchida}} (Eds.). \bibinfo{publisher}{Springer}, \bibinfo{pages}{548--563}.
\newblock
\urldef\tempurl%
\url{https://doi.org/10.1007/978-3-030-86549-8\_35}
\showDOI{\tempurl}


\bibitem[Liu et~al\mbox{.}(2020)]%
        {liu-2020-roberta}
\bibfield{author}{\bibinfo{person}{Yinhan Liu}, \bibinfo{person}{Myle Ott}, \bibinfo{person}{Naman Goyal}, \bibinfo{person}{Jingfei Du}, \bibinfo{person}{Mandar Joshi}, \bibinfo{person}{Danqi Chen}, \bibinfo{person}{Omer Levy}, \bibinfo{person}{Mike Lewis}, \bibinfo{person}{Luke Zettlemoyer}, {and} \bibinfo{person}{Veselin Stoyanov}.} \bibinfo{year}{2020}\natexlab{}.
\newblock \showarticletitle{{RoBERTa}: A Robustly Optimized {BERT} Pretraining Approach}.
\newblock \bibinfo{journal}{\emph{CoRR}} (\bibinfo{year}{2020}).
\newblock
\urldef\tempurl%
\url{https://openreview.net/forum?id=SyxS0T4tvS}
\showURL{%
\tempurl}


\bibitem[Mathew et~al\mbox{.}(2021)]%
        {mathew2021docvqa}
\bibfield{author}{\bibinfo{person}{Minesh Mathew}, \bibinfo{person}{Dimosthenis Karatzas}, {and} \bibinfo{person}{C.~V. Jawahar}.} \bibinfo{year}{2021}\natexlab{}.
\newblock \showarticletitle{Doc{VQA}: {A} Dataset for {VQA} on Document Images}. In \bibinfo{booktitle}{\emph{{IEEE} Winter Conference on Applications of Computer Vision, {WACV} 2021, Waikoloa, HI, USA, January 3-8, 2021}}. \bibinfo{publisher}{{IEEE}}, \bibinfo{pages}{2199--2208}.
\newblock
\urldef\tempurl%
\url{https://doi.org/10.1109/WACV48630.2021.00225}
\showDOI{\tempurl}


\bibitem[Neumann et~al\mbox{.}(2021)]%
        {neumann-etal-2021-pawls}
\bibfield{author}{\bibinfo{person}{Mark Neumann}, \bibinfo{person}{Zejiang Shen}, {and} \bibinfo{person}{Sam Skjonsberg}.} \bibinfo{year}{2021}\natexlab{}.
\newblock \showarticletitle{{PAWLS}: {PDF} Annotation With Labels and Structure}. In \bibinfo{booktitle}{\emph{Proceedings of the 59th Annual Meeting of the Association for Computational Linguistics and the 11th International Joint Conference on Natural Language Processing: System Demonstrations}}. \bibinfo{publisher}{Association for Computational Linguistics}, \bibinfo{address}{Online}, \bibinfo{pages}{258--264}.
\newblock
\urldef\tempurl%
\url{https://doi.org/10.18653/v1/2021.acl-demo.31}
\showDOI{\tempurl}


\bibitem[Park et~al\mbox{.}(2019)]%
        {park2019cord}
\bibfield{author}{\bibinfo{person}{Seunghyun Park}, \bibinfo{person}{Seung Shin}, \bibinfo{person}{Bado Lee}, \bibinfo{person}{Junyeop Lee}, \bibinfo{person}{Jaeheung Surh}, \bibinfo{person}{Minjoon Seo}, {and} \bibinfo{person}{Hwalsuk Lee}.} \bibinfo{year}{2019}\natexlab{}.
\newblock \showarticletitle{{CORD}: a consolidated receipt dataset for post-OCR parsing}. In \bibinfo{booktitle}{\emph{Workshop on Document Intelligence at NeurIPS 2019}}.
\newblock


\bibitem[Peng et~al\mbox{.}(2022)]%
        {peng2022ernielayout}
\bibfield{author}{\bibinfo{person}{Qiming Peng}, \bibinfo{person}{Yinxu Pan}, \bibinfo{person}{Wenjin Wang}, \bibinfo{person}{Bin Luo}, \bibinfo{person}{Zhenyu Zhang}, \bibinfo{person}{Zhengjie Huang}, \bibinfo{person}{Yuhui Cao}, \bibinfo{person}{Weichong Yin}, \bibinfo{person}{Yongfeng Chen}, \bibinfo{person}{Yin Zhang}, \bibinfo{person}{Shikun Feng}, \bibinfo{person}{Yu Sun}, \bibinfo{person}{Hao Tian}, \bibinfo{person}{Hua Wu}, {and} \bibinfo{person}{Haifeng Wang}.} \bibinfo{year}{2022}\natexlab{}.
\newblock \showarticletitle{{ERNIE}-{L}ayout: Layout Knowledge Enhanced Pre-training for Visually-rich Document Understanding}. In \bibinfo{booktitle}{\emph{Findings of the Association for Computational Linguistics: EMNLP 2022}}. \bibinfo{publisher}{Association for Computational Linguistics}, \bibinfo{address}{Abu Dhabi, United Arab Emirates}, \bibinfo{pages}{3744--3756}.
\newblock
\urldef\tempurl%
\url{https://aclanthology.org/2022.findings-emnlp.274}
\showURL{%
\tempurl}


\bibitem[Powalski et~al\mbox{.}(2021)]%
        {powalski2021going}
\bibfield{author}{\bibinfo{person}{Rafal Powalski}, \bibinfo{person}{Lukasz Borchmann}, \bibinfo{person}{Dawid Jurkiewicz}, \bibinfo{person}{Tomasz Dwojak}, \bibinfo{person}{Michal Pietruszka}, {and} \bibinfo{person}{Gabriela Palka}.} \bibinfo{year}{2021}\natexlab{}.
\newblock \showarticletitle{Going Full-{TILT} Boogie on Document Understanding with Text-Image-Layout Transformer}. In \bibinfo{booktitle}{\emph{16th International Conference on Document Analysis and Recognition, {ICDAR} 2021, Lausanne, Switzerland, September 5-10, 2021, Proceedings, Part {II}}} \emph{(\bibinfo{series}{Lecture Notes in Computer Science}, Vol.~\bibinfo{volume}{12822})}, \bibfield{editor}{\bibinfo{person}{Josep Llad{\'{o}}s}, \bibinfo{person}{Daniel Lopresti}, {and} \bibinfo{person}{Seiichi Uchida}} (Eds.). \bibinfo{publisher}{Springer}, \bibinfo{pages}{732--747}.
\newblock
\urldef\tempurl%
\url{https://doi.org/10.1007/978-3-030-86331-9\_47}
\showDOI{\tempurl}


\bibitem[Qi et~al\mbox{.}(2022)]%
        {qi-2022-dureadervis}
\bibfield{author}{\bibinfo{person}{Le Qi}, \bibinfo{person}{Shangwen Lv}, \bibinfo{person}{Hongyu Li}, \bibinfo{person}{Jing Liu}, \bibinfo{person}{Yu Zhang}, \bibinfo{person}{Qiaoqiao She}, \bibinfo{person}{Hua Wu}, \bibinfo{person}{Haifeng Wang}, {and} \bibinfo{person}{Ting Liu}.} \bibinfo{year}{2022}\natexlab{}.
\newblock \showarticletitle{$\textrm{DuReader}_{\textrm{vis}}$: A {C}hinese Dataset for Open-domain Document Visual Question Answering}. In \bibinfo{booktitle}{\emph{Findings of the Association for Computational Linguistics: ACL 2022}}. \bibinfo{publisher}{Association for Computational Linguistics}, \bibinfo{address}{Dublin, Ireland}, \bibinfo{pages}{1338--1351}.
\newblock
\urldef\tempurl%
\url{https://doi.org/10.18653/v1/2022.findings-acl.105}
\showDOI{\tempurl}


\bibitem[Rajpurkar et~al\mbox{.}(2016)]%
        {rajpurkar-etal-2016-squad}
\bibfield{author}{\bibinfo{person}{Pranav Rajpurkar}, \bibinfo{person}{Jian Zhang}, \bibinfo{person}{Konstantin Lopyrev}, {and} \bibinfo{person}{Percy Liang}.} \bibinfo{year}{2016}\natexlab{}.
\newblock \showarticletitle{{SQ}u{AD}: 100,000+ Questions for Machine Comprehension of Text}. In \bibinfo{booktitle}{\emph{Proceedings of the 2016 Conference on Empirical Methods in Natural Language Processing}}. \bibinfo{publisher}{Association for Computational Linguistics}, \bibinfo{address}{Austin, Texas}, \bibinfo{pages}{2383--2392}.
\newblock
\urldef\tempurl%
\url{https://doi.org/10.18653/v1/D16-1264}
\showDOI{\tempurl}


\bibitem[Simsa et~al\mbox{.}(2023)]%
        {simsa2023docile}
\bibfield{author}{\bibinfo{person}{Step{\'{a}}n Simsa}, \bibinfo{person}{Milan Sulc}, \bibinfo{person}{Michal Uric{\'{a}}r}, \bibinfo{person}{Yash Patel}, \bibinfo{person}{Ahmed Hamdi}, \bibinfo{person}{Matej Koci{\'{a}}n}, \bibinfo{person}{Maty{\'{a}}s Skalick{\'{y}}}, \bibinfo{person}{Jir{\'{\i}} Matas}, \bibinfo{person}{Antoine Doucet}, \bibinfo{person}{Micka{\"{e}}l Coustaty}, {and} \bibinfo{person}{Dimosthenis Karatzas}.} \bibinfo{year}{2023}\natexlab{}.
\newblock \showarticletitle{Doc{ILE} Benchmark for Document Information Localization and Extraction}.
\newblock \bibinfo{journal}{\emph{CoRR}}  \bibinfo{volume}{abs/2302.05658} (\bibinfo{year}{2023}).
\newblock
\urldef\tempurl%
\url{https://doi.org/10.48550/arXiv.2302.05658}
\showDOI{\tempurl}
\showeprint[arXiv]{2302.05658}


\bibitem[Stanis{\l}awek et~al\mbox{.}(2021)]%
        {stanislawek-2021-kleister}
\bibfield{author}{\bibinfo{person}{Tomasz Stanis{\l}awek}, \bibinfo{person}{Filip Grali{\'{n}}ski}, \bibinfo{person}{Anna Wr{\'o}blewska}, \bibinfo{person}{Dawid Lipi{\'{n}}ski}, \bibinfo{person}{Agnieszka Kaliska}, \bibinfo{person}{Paulina Rosalska}, \bibinfo{person}{Bartosz Topolski}, {and} \bibinfo{person}{Przemys{\l}aw Biecek}.} \bibinfo{year}{2021}\natexlab{}.
\newblock \showarticletitle{Kleister: {K}ey Information Extraction Datasets Involving Long Documents with Complex Layouts}. In \bibinfo{booktitle}{\emph{Document Analysis and Recognition -- ICDAR 2021}}, \bibfield{editor}{\bibinfo{person}{Josep Llad{\'o}s}, \bibinfo{person}{Daniel Lopresti}, {and} \bibinfo{person}{Seiichi Uchida}} (Eds.). \bibinfo{publisher}{Springer International Publishing}, \bibinfo{address}{Cham}, \bibinfo{pages}{564--579}.
\newblock
\showISBNx{978-3-030-86549-8}


\bibitem[Tang et~al\mbox{.}(2023)]%
        {tang2023unifying}
\bibfield{author}{\bibinfo{person}{Zineng Tang}, \bibinfo{person}{Ziyi Yang}, \bibinfo{person}{Guoxin Wang}, \bibinfo{person}{Yuwei Fang}, \bibinfo{person}{Yang Liu}, \bibinfo{person}{Chenguang Zhu}, \bibinfo{person}{Michael Zeng}, \bibinfo{person}{Cha Zhang}, {and} \bibinfo{person}{Mohit Bansal}.} \bibinfo{year}{2023}\natexlab{}.
\newblock \showarticletitle{Unifying Vision, Text, and Layout for Universal Document Processing}. In \bibinfo{booktitle}{\emph{{IEEE/CVF} Conference on Computer Vision and Pattern Recognition, {CVPR} 2023, Vancouver, BC, Canada, June 17-24, 2023}}. \bibinfo{publisher}{{IEEE}}, \bibinfo{pages}{19254--19264}.
\newblock
\urldef\tempurl%
\url{https://doi.org/10.1109/CVPR52729.2023.01845}
\showDOI{\tempurl}


\bibitem[Touvron et~al\mbox{.}(2023)]%
        {touvron2023llama}
\bibfield{author}{\bibinfo{person}{Hugo Touvron}, \bibinfo{person}{Louis Martin}, \bibinfo{person}{Kevin Stone}, \bibinfo{person}{Peter Albert}, \bibinfo{person}{Amjad Almahairi}, \bibinfo{person}{Yasmine Babaei}, \bibinfo{person}{Nikolay Bashlykov}, \bibinfo{person}{Soumya Batra}, \bibinfo{person}{Prajjwal Bhargava}, \bibinfo{person}{Shruti Bhosale}, \bibinfo{person}{Dan Bikel}, \bibinfo{person}{Lukas Blecher}, \bibinfo{person}{Cristian~Canton Ferrer}, \bibinfo{person}{Moya Chen}, \bibinfo{person}{Guillem Cucurull}, \bibinfo{person}{David Esiobu}, \bibinfo{person}{Jude Fernandes}, \bibinfo{person}{Jeremy Fu}, \bibinfo{person}{Wenyin Fu}, \bibinfo{person}{Brian Fuller}, \bibinfo{person}{Cynthia Gao}, \bibinfo{person}{Vedanuj Goswami}, \bibinfo{person}{Naman Goyal}, \bibinfo{person}{Anthony Hartshorn}, \bibinfo{person}{Saghar Hosseini}, \bibinfo{person}{Rui Hou}, \bibinfo{person}{Hakan Inan}, \bibinfo{person}{Marcin Kardas}, \bibinfo{person}{Viktor Kerkez}, \bibinfo{person}{Madian Khabsa},
  \bibinfo{person}{Isabel Kloumann}, \bibinfo{person}{Artem Korenev}, \bibinfo{person}{Punit~Singh Koura}, \bibinfo{person}{Marie-Anne Lachaux}, \bibinfo{person}{Thibaut Lavril}, \bibinfo{person}{Jenya Lee}, \bibinfo{person}{Diana Liskovich}, \bibinfo{person}{Yinghai Lu}, \bibinfo{person}{Yuning Mao}, \bibinfo{person}{Xavier Martinet}, \bibinfo{person}{Todor Mihaylov}, \bibinfo{person}{Pushkar Mishra}, \bibinfo{person}{Igor Molybog}, \bibinfo{person}{Yixin Nie}, \bibinfo{person}{Andrew Poulton}, \bibinfo{person}{Jeremy Reizenstein}, \bibinfo{person}{Rashi Rungta}, \bibinfo{person}{Kalyan Saladi}, \bibinfo{person}{Alan Schelten}, \bibinfo{person}{Ruan Silva}, \bibinfo{person}{Eric~Michael Smith}, \bibinfo{person}{Ranjan Subramanian}, \bibinfo{person}{Xiaoqing~Ellen Tan}, \bibinfo{person}{Binh Tang}, \bibinfo{person}{Ross Taylor}, \bibinfo{person}{Adina Williams}, \bibinfo{person}{Jian~Xiang Kuan}, \bibinfo{person}{Puxin Xu}, \bibinfo{person}{Zheng Yan}, \bibinfo{person}{Iliyan Zarov}, \bibinfo{person}{Yuchen
  Zhang}, \bibinfo{person}{Angela Fan}, \bibinfo{person}{Melanie Kambadur}, \bibinfo{person}{Sharan Narang}, \bibinfo{person}{Aurelien Rodriguez}, \bibinfo{person}{Robert Stojnic}, \bibinfo{person}{Sergey Edunov}, {and} \bibinfo{person}{Thomas Scialom}.} \bibinfo{year}{2023}\natexlab{}.
\newblock \bibinfo{title}{Llama 2: Open Foundation and Fine-Tuned Chat Models}.
\newblock
\newblock
\showeprint[arxiv]{2307.09288}~[cs.CL]


\bibitem[Vu and Nguyen(2020)]%
        {vu-2020-revising}
\bibfield{author}{\bibinfo{person}{Hieu~M. Vu} {and} \bibinfo{person}{Diep~Thi{-}Ngoc Nguyen}.} \bibinfo{year}{2020}\natexlab{}.
\newblock \showarticletitle{Revising {FUNSD} dataset for key-value detection in document images}.
\newblock \bibinfo{journal}{\emph{CoRR}}  \bibinfo{volume}{abs/2010.05322} (\bibinfo{year}{2020}).
\newblock
\showeprint[arXiv]{2010.05322}
\urldef\tempurl%
\url{https://arxiv.org/abs/2010.05322}
\showURL{%
\tempurl}


\bibitem[Wang et~al\mbox{.}(2023a)]%
        {wang2023docllm}
\bibfield{author}{\bibinfo{person}{Dongsheng Wang}, \bibinfo{person}{Natraj Raman}, \bibinfo{person}{Mathieu Sibue}, \bibinfo{person}{Zhiqiang Ma}, \bibinfo{person}{Petr Babkin}, \bibinfo{person}{Simerjot Kaur}, \bibinfo{person}{Yulong Pei}, \bibinfo{person}{Armineh Nourbakhsh}, {and} \bibinfo{person}{Xiaomo Liu}.} \bibinfo{year}{2023}\natexlab{a}.
\newblock \bibinfo{title}{{DocLLM}: {A} layout-aware generative language model for multimodal document understanding}.
\newblock
\newblock
\showeprint[arxiv]{2401.00908}~[cs.CL]


\bibitem[Wang et~al\mbox{.}(2020)]%
        {wang2020docstruct}
\bibfield{author}{\bibinfo{person}{Zilong Wang}, \bibinfo{person}{Mingjie Zhan}, \bibinfo{person}{Xuebo Liu}, {and} \bibinfo{person}{Ding Liang}.} \bibinfo{year}{2020}\natexlab{}.
\newblock \showarticletitle{{D}oc{S}truct: A Multimodal Method to Extract Hierarchy Structure in Document for General Form Understanding}. In \bibinfo{booktitle}{\emph{Findings of the Association for Computational Linguistics: EMNLP 2020}}. \bibinfo{publisher}{Association for Computational Linguistics}, \bibinfo{address}{Online}, \bibinfo{pages}{898--908}.
\newblock
\urldef\tempurl%
\url{https://doi.org/10.18653/v1/2020.findings-emnlp.80}
\showDOI{\tempurl}


\bibitem[Wang et~al\mbox{.}(2023b)]%
        {wang-2023-vrdu}
\bibfield{author}{\bibinfo{person}{Zilong Wang}, \bibinfo{person}{Yichao Zhou}, \bibinfo{person}{Wei Wei}, \bibinfo{person}{Chen-Yu Lee}, {and} \bibinfo{person}{Sandeep Tata}.} \bibinfo{year}{2023}\natexlab{b}.
\newblock \showarticletitle{{VRDU}: {A} Benchmark for Visually-Rich Document Understanding}. In \bibinfo{booktitle}{\emph{Proceedings of the 29th ACM SIGKDD Conference on Knowledge Discovery and Data Mining}} (Long Beach, CA, USA) \emph{(\bibinfo{series}{KDD '23})}. \bibinfo{publisher}{Association for Computing Machinery}, \bibinfo{address}{New York, NY, USA}, \bibinfo{pages}{5184–5193}.
\newblock
\showISBNx{9798400701030}
\urldef\tempurl%
\url{https://doi.org/10.1145/3580305.3599929}
\showDOI{\tempurl}


\bibitem[Xu et~al\mbox{.}(2020)]%
        {xu2020layoutlm}
\bibfield{author}{\bibinfo{person}{Yiheng Xu}, \bibinfo{person}{Minghao Li}, \bibinfo{person}{Lei Cui}, \bibinfo{person}{Shaohan Huang}, \bibinfo{person}{Furu Wei}, {and} \bibinfo{person}{Ming Zhou}.} \bibinfo{year}{2020}\natexlab{}.
\newblock \showarticletitle{{L}ayout{LM}: Pre-training of Text and Layout for Document Image Understanding}. In \bibinfo{booktitle}{\emph{{KDD} '20: The 26th {ACM} {SIGKDD} Conference on Knowledge Discovery and Data Mining, Virtual Event, CA, USA, August 23-27, 2020}}, \bibfield{editor}{\bibinfo{person}{Rajesh Gupta}, \bibinfo{person}{Yan Liu}, \bibinfo{person}{Jiliang Tang}, {and} \bibinfo{person}{B.~Aditya Prakash}} (Eds.). \bibinfo{publisher}{{ACM}}, \bibinfo{pages}{1192--1200}.
\newblock
\urldef\tempurl%
\url{https://doi.org/10.1145/3394486.3403172}
\showDOI{\tempurl}


\bibitem[Xu et~al\mbox{.}(2022)]%
        {xu-etal-2022-xfund}
\bibfield{author}{\bibinfo{person}{Yiheng Xu}, \bibinfo{person}{Tengchao Lv}, \bibinfo{person}{Lei Cui}, \bibinfo{person}{Guoxin Wang}, \bibinfo{person}{Yijuan Lu}, \bibinfo{person}{Dinei Florencio}, \bibinfo{person}{Cha Zhang}, {and} \bibinfo{person}{Furu Wei}.} \bibinfo{year}{2022}\natexlab{}.
\newblock \showarticletitle{{XFUND}: A Benchmark Dataset for Multilingual Visually Rich Form Understanding}. In \bibinfo{booktitle}{\emph{Findings of the Association for Computational Linguistics: ACL 2022}}. \bibinfo{publisher}{Association for Computational Linguistics}, \bibinfo{address}{Dublin, Ireland}, \bibinfo{pages}{3214--3224}.
\newblock
\urldef\tempurl%
\url{https://doi.org/10.18653/v1/2022.findings-acl.253}
\showDOI{\tempurl}


\bibitem[Xu et~al\mbox{.}(2021)]%
        {xu2021layoutlmv2}
\bibfield{author}{\bibinfo{person}{Yang Xu}, \bibinfo{person}{Yiheng Xu}, \bibinfo{person}{Tengchao Lv}, \bibinfo{person}{Lei Cui}, \bibinfo{person}{Furu Wei}, \bibinfo{person}{Guoxin Wang}, \bibinfo{person}{Yijuan Lu}, \bibinfo{person}{Dinei A.~F. Flor{\^{e}}ncio}, \bibinfo{person}{Cha Zhang}, \bibinfo{person}{Wanxiang Che}, \bibinfo{person}{Min Zhang}, {and} \bibinfo{person}{Lidong Zhou}.} \bibinfo{year}{2021}\natexlab{}.
\newblock \showarticletitle{{L}ayout{LM}v2: Multi-modal Pre-training for Visually-rich Document Understanding}. In \bibinfo{booktitle}{\emph{Proceedings of the 59th Annual Meeting of the Association for Computational Linguistics and the 11th International Joint Conference on Natural Language Processing, {ACL/IJCNLP} 2021, (Volume 1: Long Papers), Virtual Event, August 1-6, 2021}}, \bibfield{editor}{\bibinfo{person}{Chengqing Zong}, \bibinfo{person}{Fei Xia}, \bibinfo{person}{Wenjie Li}, {and} \bibinfo{person}{Roberto Navigli}} (Eds.). \bibinfo{publisher}{Association for Computational Linguistics}, \bibinfo{pages}{2579--2591}.
\newblock
\urldef\tempurl%
\url{https://doi.org/10.18653/v1/2021.acl-long.201}
\showDOI{\tempurl}


\bibitem[Yang et~al\mbox{.}(2015)]%
        {yang-etal-2015-wikiqa}
\bibfield{author}{\bibinfo{person}{Yi Yang}, \bibinfo{person}{Wen-tau Yih}, {and} \bibinfo{person}{Christopher Meek}.} \bibinfo{year}{2015}\natexlab{}.
\newblock \showarticletitle{{W}iki{QA}: A Challenge Dataset for Open-Domain Question Answering}. In \bibinfo{booktitle}{\emph{Proceedings of the 2015 Conference on Empirical Methods in Natural Language Processing}}. \bibinfo{publisher}{Association for Computational Linguistics}, \bibinfo{address}{Lisbon, Portugal}, \bibinfo{pages}{2013--2018}.
\newblock
\urldef\tempurl%
\url{https://doi.org/10.18653/v1/D15-1237}
\showDOI{\tempurl}


\bibitem[Yang et~al\mbox{.}(2018)]%
        {yang-etal-2018-hotpotqa}
\bibfield{author}{\bibinfo{person}{Zhilin Yang}, \bibinfo{person}{Peng Qi}, \bibinfo{person}{Saizheng Zhang}, \bibinfo{person}{Yoshua Bengio}, \bibinfo{person}{William Cohen}, \bibinfo{person}{Ruslan Salakhutdinov}, {and} \bibinfo{person}{Christopher~D. Manning}.} \bibinfo{year}{2018}\natexlab{}.
\newblock \showarticletitle{{H}otpot{QA}: A Dataset for Diverse, Explainable Multi-hop Question Answering}. In \bibinfo{booktitle}{\emph{Proceedings of the 2018 Conference on Empirical Methods in Natural Language Processing}}. \bibinfo{publisher}{Association for Computational Linguistics}, \bibinfo{address}{Brussels, Belgium}, \bibinfo{pages}{2369--2380}.
\newblock
\urldef\tempurl%
\url{https://doi.org/10.18653/v1/D18-1259}
\showDOI{\tempurl}


\bibitem[Ye et~al\mbox{.}(2023)]%
        {ye2023mplugdocowl}
\bibfield{author}{\bibinfo{person}{Jiabo Ye}, \bibinfo{person}{Anwen Hu}, \bibinfo{person}{Haiyang Xu}, \bibinfo{person}{Qinghao Ye}, \bibinfo{person}{Ming Yan}, \bibinfo{person}{Yuhao Dan}, \bibinfo{person}{Chenlin Zhao}, \bibinfo{person}{Guohai Xu}, \bibinfo{person}{Chenliang Li}, \bibinfo{person}{Junfeng Tian}, \bibinfo{person}{Qian Qi}, \bibinfo{person}{Ji Zhang}, {and} \bibinfo{person}{Fei Huang}.} \bibinfo{year}{2023}\natexlab{}.
\newblock \bibinfo{title}{{mPLUG-DocOwl}: {M}odularized Multimodal Large Language Model for Document Understanding}.
\newblock
\newblock
\showeprint[arxiv]{2307.02499}~[cs.CL]


\bibitem[Zhang et~al\mbox{.}(2020)]%
        {zhang2021trie}
\bibfield{author}{\bibinfo{person}{Peng Zhang}, \bibinfo{person}{Yunlu Xu}, \bibinfo{person}{Zhanzhan Cheng}, \bibinfo{person}{Shiliang Pu}, \bibinfo{person}{Jing Lu}, \bibinfo{person}{Liang Qiao}, \bibinfo{person}{Yi Niu}, {and} \bibinfo{person}{Fei Wu}.} \bibinfo{year}{2020}\natexlab{}.
\newblock \showarticletitle{{TRIE:} {E}nd-to-End Text Reading and Information Extraction for Document Understanding}. In \bibinfo{booktitle}{\emph{{MM} '20: The 28th {ACM} International Conference on Multimedia, Virtual Event / Seattle, WA, USA, October 12-16, 2020}}, \bibfield{editor}{\bibinfo{person}{Chang~Wen Chen}, \bibinfo{person}{Rita Cucchiara}, \bibinfo{person}{Xian{-}Sheng Hua}, \bibinfo{person}{Guo{-}Jun Qi}, \bibinfo{person}{Elisa Ricci}, \bibinfo{person}{Zhengyou Zhang}, {and} \bibinfo{person}{Roger Zimmermann}} (Eds.). \bibinfo{publisher}{{ACM}}, \bibinfo{pages}{1413--1422}.
\newblock
\urldef\tempurl%
\url{https://doi.org/10.1145/3394171.3413900}
\showDOI{\tempurl}


\bibitem[Zhang et~al\mbox{.}(2022)]%
        {zhang2022multimodal}
\bibfield{author}{\bibinfo{person}{Zhenrong Zhang}, \bibinfo{person}{Jiefeng Ma}, \bibinfo{person}{Jun Du}, \bibinfo{person}{Licheng Wang}, {and} \bibinfo{person}{Jianshu Zhang}.} \bibinfo{year}{2022}\natexlab{}.
\newblock \showarticletitle{Multimodal Pre-training Based on Graph Attention Network for Document Understanding}.
\newblock \bibinfo{journal}{\emph{CoRR}}  \bibinfo{volume}{abs/2203.13530} (\bibinfo{year}{2022}).
\newblock
\urldef\tempurl%
\url{https://doi.org/10.48550/arXiv.2203.13530}
\showDOI{\tempurl}
\showeprint[arXiv]{2203.13530}


\bibitem[Zmigrod et~al\mbox{.}(2024)]%
        {zmigrod-2024-treeform}
\bibfield{author}{\bibinfo{person}{Ran Zmigrod}, \bibinfo{person}{Zhiqiang Ma}, \bibinfo{person}{Armineh Nourbakhsh}, {and} \bibinfo{person}{Sameena Shah}.} \bibinfo{year}{2024}\natexlab{}.
\newblock \showarticletitle{{T}ree{F}orm: End-to-end Annotation and Evaluation for Form Document Parsing}. In \bibinfo{booktitle}{\emph{Proceedings of The 18th Linguistic Annotation Workshop (LAW-XVIII)}}, \bibfield{editor}{\bibinfo{person}{Sophie Henning} {and} \bibinfo{person}{Manfred Stede}} (Eds.). \bibinfo{publisher}{Association for Computational Linguistics}, \bibinfo{address}{St. Julians, Malta}, \bibinfo{pages}{1--11}.
\newblock
\urldef\tempurl%
\url{https://aclanthology.org/2024.law-1.1}
\showURL{%
\tempurl}


\end{thebibliography}

\clearpage
\appendix

\section{\datasetName Annotation Instructions}\label{app:annot}
In this section, we provide a more detailed description of the instructions received by annotators for the DC and KEE annotation tasks. 
For both tasks, annotators first annotated their assigned documents using the instructions provided below.
Then, a validator was assigned to check these annotations using the same instructions.
Any major disagreements that the validator and annotator were not able to resolve with the help of a third annotator were discarded.

\subsection{Document Classification}
Annotators were instructed to pick a document class using these ordered instructions.

\begin{enumerate}
    \item If the document title contains the word ``detail'', ``business'', ``entity'', or ``search'', classify the document as \emph{Business Entity Details}.
    
    \item If the document title contains the word ``annual'', ``biennial'', ``periodic'', etc., or contains a year (e.g., 2007), classify the document as \emph{Periodic Report}.
    
    \item If the document title contains the word ``amend'', ``update'', or ``change'', classify the document as \emph{Amendment Document}.
    
    \item If the document title contains the word ``application'', ``article'', or ``reservation', classify the document as \emph{Article or Application}.
    
    \item If the document title contains the word ``certificate'', ``statement'', ``affidavit'', ``report'', ``confirmation'', ``notice'', or ``receipt'', classify the document as \emph{Certificate or Statement}. Note that an ``Application for a Certificate'' should be classified as \emph{Article or Application} by the previous instruction.

    \item If there is no title, examine the format and content; if it seems descriptive of a business, classify the document as \emph{Business Entity Details}.
    
    \item If none of the above rules hold, do not label this document. 
\end{enumerate}

\subsection{Key Entity Extraction}
For the KEE task, annotators utilised an annotation tool that allowed them to create labelled bounding boxes where the labels available are given in \cref{tab:app-entity-stats} (an additional \texttt{is\_key} label was annotated but not included in this version of \datasetName).
Annotators were asked to abide by the following annotation instructions.
\begin{enumerate}
    \item For each meaningful value in the document, check whether the value relates to any of the super categories (given in \cref{tab:entity-stats}). If no super category is identified but you are sure this is a meaningful value, select the \texttt{OTHER} category. Please see below for examples for some of the super categories.
    \begin{itemize}
        \item Business Entity (\texttt{ENT}): Corporation, business, trade, etc.
        \item Government Official (\texttt{GO}): State secretary, mayor, etc.
        \item Key Personal (\texttt{KP}): Director, vice president, treasurer, etc.
    \end{itemize}

    \item Select from the fine-grained labels (given in \cref{tab:app-entity-stats}) of the category the appropriate label for the value. If the value does not have an appropriate label, omit the annotation.

    \item Create a bounding box around the value tokens. This will select all OCR tokens that are in or lay on the bounding. If this selection is not accurate, you may also turn off the OCR selection tool and draw a free form bounding box. \emph{Note: For this version of the dataset, we only include bounding boxes that use the OCR selection tool.}

    \item If the value has an associated key, select the \texttt{is\_key} label and create a bounding box as in the previous step. \emph{Note: For this version of the dataset, we did not include the \texttt{is\_key} entities.}

    \item Only create an annotation if you are sure that the value is meaningful and you have chosen the correct label.
\end{enumerate}

All annotators first annotated ten practice documents for which they received feedback before they began annotating the dataset documents.

\section{Data Statistics}\label{app:data}
In this section, we provide further breakdowns on the number of occurrences for each annotation type in \datasetName.
Firstly, \cref{tab:app-entity-stats} gives details on the occurrences of 
key entity labels in the dataset.
Secondly, \cref{tab:state-stats} provides the same breakdown for the number of documents originating from each US state.

\begin{table*}[t!]
    \centering
    \small
    \resizebox{.99\textwidth}{!}{
\begin{tabular}{lrrrr lrrrr lrrrr}
\bf Entity Label & \bf Total & \bf Train & \bf Val & \bf Test & 
\bf Entity Label & \bf Total & \bf Train & \bf Val & \bf Test &
\bf Entity Label & \bf Total & \bf Train & \bf Val & \bf Test \\
\midrule
{\footnotesize \texttt{AGT\_adrs\_city}} & $1174$ & $820$ & $113$ & $241$ & {\footnotesize \texttt{ENT\_residency}} & $152$ & $106$ & $19$ & $27$ & {\footnotesize \texttt{GO\_fax}} & $39$ & $28$ & $2$ & $9$ \\ 
{\footnotesize \texttt{AGT\_adrs\_country}} & $240$ & $162$ & $24$ & $54$ & {\footnotesize \texttt{ENT\_shares\_auth}} & $50$ & $43$ & $3$ & $4$ & {\footnotesize \texttt{GO\_telephone}} & $262$ & $182$ & $23$ & $57$ \\ 
{\footnotesize \texttt{AGT\_adrs\_state}} & $1150$ & $806$ & $112$ & $232$ & {\footnotesize \texttt{ENT\_shares\_issued}} & $50$ & $33$ & $4$ & $13$ & {\footnotesize \texttt{GO\_website}} & $212$ & $146$ & $23$ & $43$ \\ 
{\footnotesize \texttt{AGT\_adrs\_street}} & $1146$ & $802$ & $109$ & $235$ & {\footnotesize \texttt{ENT\_status}} & $806$ & $552$ & $85$ & $169$ & {\footnotesize \texttt{GO\_name}} & $480$ & $360$ & $47$ & $73$ \\ 
{\footnotesize \texttt{AGT\_adrs\_zipcode}} & $1148$ & $804$ & $109$ & $235$ & {\footnotesize \texttt{ENT\_type}} & $1041$ & $727$ & $103$ & $211$ & {\footnotesize \texttt{GO\_title}} & $627$ & $462$ & $60$ & $105$ \\ 
{\footnotesize \texttt{AGT\_name}} & $1214$ & $854$ & $115$ & $245$ & {\footnotesize \texttt{FILE\_adrs\_city}} & $70$ & $50$ & $9$ & $11$ & {\footnotesize \texttt{KP\_adrs\_city}} & $1413$ & $972$ & $144$ & $297$ \\ 
{\footnotesize \texttt{ENT\_NAICS}} & $107$ & $70$ & $13$ & $24$ & {\footnotesize \texttt{FILE\_adrs\_state}} & $114$ & $81$ & $13$ & $20$ & {\footnotesize \texttt{KP\_adrs\_country}} & $490$ & $350$ & $37$ & $103$ \\ 
{\footnotesize \texttt{ENT\_adrs\_city}} & $1552$ & $1083$ & $142$ & $327$ & {\footnotesize \texttt{FILE\_adrs\_street}} & $71$ & $50$ & $9$ & $12$ & {\footnotesize \texttt{KP\_adrs\_state}} & $1374$ & $953$ & $130$ & $291$ \\ 
{\footnotesize \texttt{ENT\_adrs\_country}} & $377$ & $253$ & $44$ & $80$ & {\footnotesize \texttt{FILE\_adrs\_zipcode}} & $71$ & $50$ & $9$ & $12$ & {\footnotesize \texttt{KP\_adrs\_street}} & $1488$ & $1026$ & $141$ & $321$ \\ 
{\footnotesize \texttt{ENT\_adrs\_state}} & $1500$ & $1046$ & $140$ & $314$ & {\footnotesize \texttt{FILE\_date}} & $907$ & $633$ & $90$ & $184$ & {\footnotesize \texttt{KP\_adrs\_zipcode}} & $1383$ & $958$ & $134$ & $291$ \\ 
{\footnotesize \texttt{ENT\_adrs\_street}} & $1485$ & $1050$ & $137$ & $298$ & {\footnotesize \texttt{FILE\_due\_date}} & $235$ & $163$ & $29$ & $43$ & {\footnotesize \texttt{KP\_name}} & $1934$ & $1337$ & $171$ & $426$ \\ 
{\footnotesize \texttt{ENT\_adrs\_zipcode}} & $1450$ & $1010$ & $135$ & $305$ & {\footnotesize \texttt{FILE\_eff\_date}} & $155$ & $115$ & $15$ & $25$ & {\footnotesize \texttt{KP\_shares\_owned}} & $78$ & $58$ & $12$ & $8$ \\ 
{\footnotesize \texttt{ENT\_alt\_name}} & $29$ & $23$ & $1$ & $5$ & {\footnotesize \texttt{FILE\_exp\_date}} & $48$ & $40$ & $2$ & $6$ & {\footnotesize \texttt{KP\_title}} & $1685$ & $1199$ & $137$ & $349$ \\ 
{\footnotesize \texttt{ENT\_am\_adrs\_city}} & $21$ & $19$ & $1$ & $1$ & {\footnotesize \texttt{FILE\_fee}} & $398$ & $284$ & $44$ & $70$ & {\footnotesize \texttt{OTHER\_unknown}} & $522$ & $404$ & $24$ & $94$ \\ 
{\footnotesize \texttt{ENT\_am\_adrs\_state}} & $21$ & $19$ & $1$ & $1$ & {\footnotesize \texttt{FILE\_name}} & $927$ & $660$ & $77$ & $190$ & {\footnotesize \texttt{OTHER\_adrs}} & $95$ & $64$ & $13$ & $18$ \\ 
{\footnotesize \texttt{ENT\_am\_adrs\_street}} & $23$ & $21$ & $1$ & $1$ & {\footnotesize \texttt{FILE\_number}} & $494$ & $345$ & $47$ & $102$ & {\footnotesize \texttt{OTHER\_date\_time}} & $185$ & $142$ & $9$ & $34$ \\ 
{\footnotesize \texttt{ENT\_am\_adrs\_zipcode}} & $20$ & $18$ & $1$ & $1$ & {\footnotesize \texttt{FILE\_state}} & $300$ & $214$ & $38$ & $48$ & {\footnotesize \texttt{OTHER\_name}} & $37$ & $28$ & $2$ & $7$ \\ 
{\footnotesize \texttt{ENT\_am\_name}} & $16$ & $13$ & $2$ & $1$ & {\footnotesize \texttt{FILE\_type}} & $238$ & $155$ & $28$ & $55$ & {\footnotesize \texttt{SIG\_GO\_date}} & $19$ & $16$ & $1$ & $2$ \\ 
{\footnotesize \texttt{ENT\_cob}} & $295$ & $200$ & $30$ & $65$ & {\footnotesize \texttt{GO\_adrs\_city}} & $344$ & $247$ & $30$ & $67$ & {\footnotesize \texttt{SIG\_GO\_name}} & $29$ & $23$ & $1$ & $5$ \\ 
{\footnotesize \texttt{ENT\_formation\_date}} & $704$ & $487$ & $69$ & $148$ & {\footnotesize \texttt{GO\_adrs\_state}} & $343$ & $245$ & $30$ & $68$ & {\footnotesize \texttt{SIG\_GO\_title}} & $45$ & $35$ & $4$ & $6$ \\ 
{\footnotesize \texttt{ENT\_jurisdiction}} & $863$ & $600$ & $84$ & $179$ & {\footnotesize \texttt{GO\_adrs\_street}} & $328$ & $235$ & $27$ & $66$ & {\footnotesize \texttt{SIG\_KP\_date}} & $317$ & $227$ & $22$ & $68$ \\ 
{\footnotesize \texttt{ENT\_name}} & $1890$ & $1330$ & $184$ & $376$ & {\footnotesize \texttt{GO\_adrs\_zipcode}} & $342$ & $245$ & $30$ & $67$ & {\footnotesize \texttt{SIG\_KP\_name}} & $488$ & $343$ & $37$ & $108$ \\ 
{\footnotesize \texttt{ENT\_number}} & $1432$ & $1000$ & $140$ & $292$ & {\footnotesize \texttt{GO\_email}} & $69$ & $47$ & $8$ & $14$ & {\footnotesize \texttt{SIG\_KP\_title}} & $294$ & $206$ & $28$ & $60$ \\ 

\end{tabular}
}
\caption{Number of occurrences in the train, validation, and test splits of \datasetName for each key entity label.}
\label{tab:app-entity-stats}
\end{table*}

\begin{table*}[t]
    \centering
    \small
\begin{tabular}{llrrrr p{0.25cm} llrrrr}
\bf State & \bf State Abb. & \bf Total & \bf Train & \bf Val & \bf Test & & \bf State & \bf State Abb. & \bf Total & \bf Train & \bf Val & \bf Test \\ \midrule
Alabama & AL & $40$ & $30$ & $4$ & $6$ & & New Hampshire & NH & $40$ & $29$ & $4$ & $7$ \\
Alaska & AK & $68$ & $47$ & $8$ & $13$ & & New Jersey & NJ & $36$ & $25$ & $0$ & $11$ \\
Arizona & AZ & $78$ & $52$ & $6$ & $20$ & & New Mexico & NM & $19$ & $12$ & $3$ & $4$ \\
Arkansas & AR & $46$ & $32$ & $6$ & $8$ & & New York & NY & $18$ & $12$ & $4$ & $2$ \\
California & CA & $25$ & $19$ & $2$ & $4$ & & North Carolina & NC & $122$ & $92$ & $9$ & $21$ \\
Connecticut & CT & $18$ & $17$ & $1$ & $0$ & & North Dakota & ND & $10$ & $9$ & $0$ & $1$ \\
Delaware & DE & $12$ & $11$ & $1$ & $0$ & & Ohio & OH & $11$ & $8$ & $0$ & $3$ \\
Florida & FL & $34$ & $24$ & $1$ & $9$ & & Oklahoma & OK & $30$ & $22$ & $2$ & $6$ \\
Georgia & GA & $58$ & $41$ & $6$ & $11$ & & Oregon & OR & $29$ & $21$ & $2$ & $6$ \\
Hawaii & HI & $35$ & $25$ & $3$ & $7$ & & Pennsylvania & PA & $53$ & $35$ & $5$ & $13$ \\
Idaho & ID & $30$ & $19$ & $3$ & $8$ & & Puerto Rico & PR & $20$ & $14$ & $3$ & $3$ \\
Iowa & IA & $96$ & $72$ & $9$ & $15$ & & Rhode Island & RI & $26$ & $21$ & $1$ & $4$ \\
Kansas & KS & $35$ & $22$ & $3$ & $10$ & & South Dakota & SD & $88$ & $61$ & $7$ & $20$ \\
Kentucky & KY & $65$ & $47$ & $7$ & $11$ & & Tennessee & TN & $24$ & $14$ & $4$ & $6$ \\
Maryland & MD & $23$ & $19$ & $1$ & $3$ & & Utah & UT & $9$ & $3$ & $2$ & $4$ \\
Massachusetts & MA & $32$ & $25$ & $1$ & $6$ & & Vermont & VT & $19$ & $12$ & $3$ & $4$ \\
Minnesota & MN & $20$ & $11$ & $4$ & $5$ & & Virginia & VA & $35$ & $23$ & $6$ & $6$ \\
Missouri & MO & $50$ & $33$ & $9$ & $8$ & & Washington & WA & $40$ & $20$ & $8$ & $12$ \\
Montana & MT & $20$ & $15$ & $1$ & $4$ & & West Virginia & WV & $10$ & $8$ & $0$ & $2$ \\
Nebraska & NE & $20$ & $12$ & $4$ & $4$ & & Wisconsin & WI & $20$ & $15$ & $4$ & $1$ \\
Nevada & NV & $40$ & $24$ & $4$ & $12$ & & Wyoming & WY & $161$ & $119$ & $10$ & $32$
\end{tabular}
\caption{Number of occurrences in the train, validation, and test splits of \datasetName for each US state. Note that Illinois, Indiana, Louisiana, Maine, Mississippi, and Texas have no documents as they were blocked by a paywall. Furthermore, documents from Colorado and Michigan were not available for the distribution purposes of \datasetName.}
    \label{tab:state-stats}
\end{table*}

\end{document}